\documentclass[journal, twocolumn]{IEEEtran}
\usepackage[colorlinks,
linkcolor=blue,
anchorcolor=blue,
citecolor=blue]{hyperref}
\usepackage[english]{babel}
\usepackage[utf8x]{inputenc}
\usepackage[T1]{fontenc}
\usepackage{graphicx}
\usepackage{subfigure}
\usepackage{minipage-marginpar}
\usepackage{amsmath}
\usepackage{amssymb}
\usepackage{cases}
\usepackage{times} 
\usepackage[font=footnotesize,labelfont=bf]{caption}
\usepackage{caption}
\usepackage{algorithm}
\usepackage{algorithmic}
\usepackage{color}
\usepackage{multirow}
\usepackage{algorithm, algorithmic}

\makeatletter
\newcommand\figcaption{\def\@captype{figure}\caption}
\newcommand\tabcaption{\def\@captype{table}\caption}
\makeatother
\usepackage{times}
\usepackage{mathptmx}
\usepackage{setspace}

\usepackage{array} 
\newcolumntype{C}[1]{>{\centering}p{#1}}
\usepackage{multicol}  
\usepackage{multirow}

\begin{document}
	
	\title{Lightweight machine unlearning in neural network}
	\author{\sffamily Kongyang Chen$^1$, \sffamily Yiwen Wang$^2$, \sffamily Yao Huang$^3$}
	
	\IEEEtitleabstractindextext{
		\begin{abstract}
			
			In recent years, machine learning neural network has penetrated deeply into people's life. As the price of convenience, people's private information also has the risk of disclosure. The "right to be forgotten" was introduced in a timely manner, stipulating that individuals have the right to withdraw their consent from personal information processing activities based on their consent. To solve this problem, machine unlearning is proposed, which allows the model to erase all memory of private information. Previous studies, including retraining and incremental learning to update models, often take up extra storage space or are difficult to apply to neural networks. Our method only needs to make a small perturbation of the weight of the target model and make it iterate in the direction of the model trained with the remaining data subset until the contribution of the unlearning data to the model is completely eliminated. In this paper, experiments on five datasets prove the effectiveness of our method for machine unlearning, and our method is 15 times faster than retraining.
			
		\end{abstract}
	}
	\maketitle
	\IEEEdisplaynontitleabstractindextext
	\IEEEpeerreviewmaketitle

	\section{Introduction}
	
	In recent years, deep learning has become a hot topic. Autonomous driving, face recognition, medical evaluation and other benefits have brought convenience to people. In order to provide better services, more expensive and high-performance deep learning models emerge one after another. Service providers need to collect a lot of data to provide the basis for these deep learning models. In order to obtain more convenience, people will continue to provide personal information for service providers to use. However, while enjoying the convenience of personal information exchange, the system can calculate and obtain more data after collecting these original information, by changing the format of photos, intercepting important parts, and extracting movie scores from movie reviews, the original information is deeply aggregated into statistical information, deriving more data for system construction.  However, the purchase record may reveal the user's savings, and the medical history mark may reveal the user's family genetic history. Our private information flows out in potential ways, in many forms and in many places we don't understand. Fortunately, more and more people are aware of the risk of privacy breaches, and users want the system to "forget" the data it has ever submitted and all its participating parts. Thus "machine unlearning" was proposed and applied.

	Differential privacy \cite{2009Differentially} protects users' privacy to some extent. It makes use of noise to make it impossible to distinguish whether a single sample exists in the dataset. The contribution of each training sample to the model is guaranteed to be limited in scope. However, different from machine unlearning, differential privacy protects each sample without distinction. Therefore, contribution to the model cannot be completely eliminated, otherwise the model cannot learn. Considering that there is a user need to completely eliminate the impact of personal data on the model, machine unlearning requires not only deleting user data from the database, but also eliminating the data's contribution to the model completely. It is easy to imagine, then, that by removing the relevant data from the training sample and retraining the model, the influence of this data could be completely eliminated. But such a direct approach would come at a huge cost in time. At present, the main methods of machine unlearning focus on retraining the model and updating the model through computable transformation \cite{2015Towards}, which are limited and difficult to be applied in the neural network framework.

	Our aim is to provide a lightweight and convenient method that costs relatively little in time and space and can achieve almost no loss of model unlearning accuracy. So we don't want to spend a lot of time retraining. Instead, based on the model before unlearning, we make a simple correction, and add a small disturbance to its weight, so that the model after change does not have the influence of data to be forgotten. Therefore, a reference direction should be determined before correction so that the model before unlearning can approach this direction. Therefore, in our method, we need a reference model to improve the direction of weight change of the target model. The selection of the reference model must not contain the information of the data to be forgotten, and the weight distribution of the reference model should be as similar as possible to the weight distribution of the target model, so that the adjustment of the forgotten target model will not affect its own performance in a small range. To be specific, the retrained model would be the best choice for the reference model if it didn't consume a lot of time. Therefore, the selection of reference model should be based on the retrained model and try to be similar to it. When unlearning, We input the data to be forgotten into the reference model to obtain the output distribution as the standard. Input the data to be forgotten into the target model to get a distribution, then iterated it to the standard, and finally the two distributions are the same, so Change the weight of the target model and complete the change of the weight of the target model to achieve the purpose of unlearning .The contributions of this paper are as follows:

	\begin{enumerate}
		\item We propose a new general machine learning method, which is compatible with all neural network architectures without the assistance of server and multiple users.
		
		\item Our method based on the model before unlearning, we make a simple correction, and add a small disturbance to its weight, so that the model after change does not have the influence of data to be forgotten. Therefore, a reference direction should be determined before correction so that the model before unlearning can approach this direction.
		
		\item Evaluating the performance of our method on five datasets.  Compared to the baseline, accuracy loss is no more than 5\% , accuracy of membership attack, backdoor attack and the time of unlearning is well below baseline.
		
	\end{enumerate}

	\section{Background and Motivation}
	Data is very important and is regarded as the oil of the new era \cite{TheEconomist}. Data is the new competitiveness. In the era of rapid explosion of information, the huge database of data information comes from all aspects of social life. Bank bill information, personal photo data and click records of social networking sites can all become user data in the background and form statistical data after in-depth aggregation. The huge amount of data provides input samples for machine learning models, which can be used to build accurate user portraits and provide various personalized services such as shopping recommendations and travel reminders.

	However, in the actual use of data, there are serious problems of information leakage and data abuse.For example, some illegal organizations collect data illegally for precision fraud, and some illegal companies design "big data killing" to maliciously raise product prices based on historical records. In order to solve these problems, some privacy computing methods have been born in recent years and are listed as one of the top ten strategic breakthrough technologies in 2020 by Gartner \cite{cearley2020top}. Typical methods are federated learning \cite{bonawitz2017practical} \cite{tran2019federated}, differential privacy, etc. They hope to help provide better machine learning models without disclosing user data. The latest research work \cite{2019Deep} shows that federated learning and other methods still pose security threats, and malicious attackers can almost completely recover local data after several iterations. This is because the machine learning model always "remembers" the local data in the training process. Techniques such as member inference attack \cite{2017Membership} can also determine whether some data is used in the model training process, which still brings new data leakage security.
	
	So we need another approach to privacy, a machine learning model that completely forgets some data, a process also known as "machine unlearning". In fact, there have been a large number of laws and regulations on machine unlearning at home and abroad, which clearly point out the need to allow users to cancel data authorization and forget the impact of data. For example, the Eu General Regulation on Data Protection \cite{voigt2017eu} specifies the "right to be forgotten" : each data subject has the right to deal with non-compliant data, and in particular, to continue to correct, erasure, and block data that is incomplete or incorrect. Personal Information Protection Law of The People's Republic of China (Draft) \cite{determann2021china} : Article 16 Individuals have the right to withdraw their consent for personal information processing activities based on their consent. Major companies also have rules about data unlearning. For example, in apple's ios14 update, the App tracking transparency feature will require an App to obtain a user's permission before tracking user data on another company's App or website.

	There has been some work on machine unlearning.Intuitively, if you want to forget some data, the easiest way is to delete the data from the input sample and retrain a machine learning model from the rest of the input sample. The disadvantage is that it consumes a lot of training time. Predecessors also put forward some work based on statistical query \cite{2015Towards}, which uses statistical query to obtain the characteristics of the dataset instead of training directly in the dataset, thus unlearning in less time than new training. However, the statistical query method is only applicable to non-adaptive machine learning models (late training does not depend on early training), and it is difficult to achieve unlearning effect on neural networks. L.Bourtoule \cite{2019Machine} divided the dataset into several parts, each part was trained into a separate sub-model and stored, and the overall model was trained through incremental learning. To forget a sample, retraining starts with the first intermediate model that contains the sample's contribution. However, this method reduces training time, but consumes a lot of storage space. Another based on the study of data lost work \cite{liu2020federated}, preserved in normal training stage of polymerization model updating parameters of each round, and then delete the memory data to training, reduce the number of iterations client training, the parameters of the model aggregation when the current client before training and keep update the parameters of the combined structure model of a service. This approach reduces training time, but also takes up more storage space because saving parameters.Due to the need to save updated parameters, the training process of the target model is modified, and the saved parameters themselves carry information to be forgotten, which theoretically cannot guarantee the complete unlearning. Obviously, previous work has shortcomings in computing time, application scenarios and so on. This paper designs a fast machine unlearning method in general scenarios.

	\section{OUR APPROACH}
	
	\subsection{OVERVIEW}
	Machine unlearning not only needs to delete the data to be forgotten, but also needs to completely eliminate the influence of the data on the model. As mentioned earlier, retrain with the remaining data is a simple and effective method, but it can be prohibitively expensive. In order to achieve efficient unlearning, we do not want to retrain on the remaining data, but make full use of the trained target model to design a new unlearning algorithm on this model, so as to achieve unlearning quickly.
	As shown in Figure \ref{fig:frame} below, assuming that complete dataset $D$ is known and the trained machine learning model is $M_{initial}$, we hope to forget part of the data $D_f$ and eliminate the influence of $D_f$ from the $M_{initial}$. Since it is slow to retrain a model from $D-D_f$, we consider taking a subset of it, let's say $D_s$, and quickly train a new model $M_0$ from $D_s$. Obviously, there is no training trace of $D_f$ data in $M_0$, but there is training trace of $D_f$ data in $M_{initial}$. If we can make $M_{initial}$ iterate in the direction of $M_0$ through incremental iterative learning to eliminate the training traces of $D_f$ data, $D_f$ unlearning from $M_{initial}$ should also be realized, that is, the final forgotten model can be obtained. In general, our method does not need to do a training process in the large-scale data, a significant increase in storage space, need not forgotten but take a subset of data after training, the target model in the direction of subset model iterative update, until forget data in both posterior distribution is almost the same, as we believe that the data has been forgotten.

	\subsection{PROBLEM DEFINATION}
	
	Let $D$ be the entire dataset, $D_f={x_i,y_i\ }(i=1)^N$$(D_f\subset D)$ is the dataset to be forgotten consisting of $N$ samples and their corresponding label $y$, and $D_r(D_r\cup D_f=D)$ is the dataset left after $D_f$ is deleted. The target model $M$, whose trainable parameter is $\theta$, is expressed as $f_{\theta}$.\\
	DEFINITION 1 MACHINE UNLEARNING:
	
	Given a dataset $D$, it is necessary to forget dataset $D_f$ and train model $f$ with $D$, whose parameter is $\theta$.The parameter of the model $h$ trained by $D_r$ is $\varepsilon$. We assume that (model $h$ obeys a posterior distribution $P$) ${h_{\varepsilon}(D_f)}\sim{P}$. If $f$ satisfies (model $f$ obeys a posterior distribution $P$ even after unlearning), machine unlearning is considered.

	\subsection{DETAIL}
	
	In our method, the core idea is to change the representation of forgotten data in the model so as to correct the weight of the model. First, in order to forget, we need to determine the direction of a weight correction. In the case of ignoring the cost, retraining with the remaining dataset $D_r$ After deleting the forgotten data $D_f$ is a simple and effective unlearning method. With this in mind, we can use the retrained model as a reference for unlearning to guide the direction of unlearning. We train a reference model $M_0$, whose trainable parameter is $\omega$, and the model is expressed as $h_{\omega}$. For time cost, a subset $D_s$$({D_s}\subset{D_r})$ is randomly selected from the remaining data as the training sample for $M_0$. The output distribution $P(\omega,x)$ of forgotten data $D_f$ on $M_0$ is calculated as a reference, and $P(\theta,x)$ represents the output distribution of $D_f$ on target model $M$.

\begin{figure*}[!h]
		\centering
		\includegraphics[width=1\textwidth]{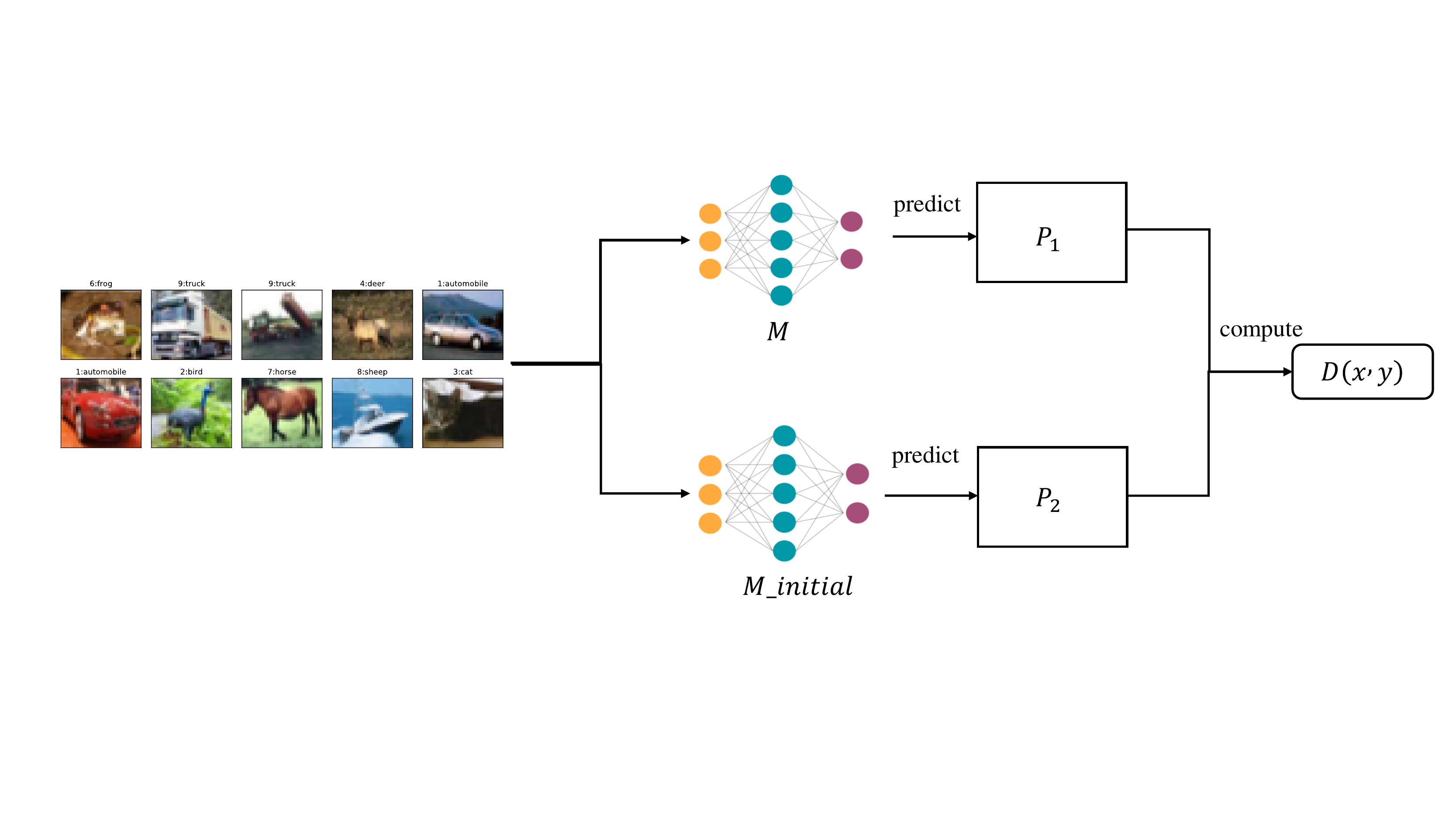}
		\caption{frame of unlearning}
		\label{fig:frame}
	\end{figure*}

	In Figure \ref{fig:frame}, the initial model before unlearning is $M_{initial}$, and $M_0$ is the reference model. The data to be forgotten is input into the reference model and the target model respectively, and the posterior distribution is calculated respectively. $M_0$ is fixed, that is, the posterior distribution of the reference model remains unchanged, and $M_{initial}$ is adjusted during iteration to make the distribution of the target model close to the distribution iteration of the reference model. At the end of iteration, the two distributions are almost the same, and $M_{initial}$ is iterated to the model $M_{final}$ after the final unlearning.

	To simulate the output distribution $P(\omega,x)$ of the forgotten data $D_f$ on the reference model $M_0$, the distance $d$ between $P(\omega,x)$ and the output distribution $P(\theta,x)$ of the forgotten data $D_f$ on $M$ is first quantified. Kullback-Leibler divergence (KL divergence)\cite{goodfellow2016nips}, a distance function commonly used in machine learning, is adopted. The simplicity of calculation helps to reduce training time.Our goal is to find a specific disturbance on the weight $\theta$ of the target model $M$ to minimize the distance $d$ between $P(\omega,x)$ and $P(\theta,x)$, thus simulating the output distribution $P(\omega,x)$ of the forgotten data $D_f$ on the reference model $M_{\theta}$. So the loss function:

	\begin{equation}
	\label{loss1}
		L_{KL}=-\sum_{x\in{D_f}}P(\omega,x)\log{P(\theta,x)}+\sum_{x\in{D_f}}P(\omega,x)\log{P(\omega,x)}
	\end{equation}
	
	In addition, in order to optimize the accuracy of the model after unlearning, it is necessary to add a penalty term. Suppose $x^\prime\in D_r$, its corresponding label is $Y$, $y^\prime=f_\theta\left(x^\prime\right)$, $H$ is the cross entropy loss \cite{barz2018detecting} between the predicted label and the real label, and the accuracy of the target model will be corrected.

	\begin{equation}
	\label{ce}
		H_{CE}\ (y\prime,Y)=-[{y\prime}\log{Y}+(1-{y\prime}\ )\ \log(\ 1-Y)\ ]
	\end{equation}

	Update loss function is:

	\begin{equation}
	\label{loss2}
		\lambda L_{KL}\ (P(\omega,x),P(\theta,x))+(1-\lambda)\ H_{CE}\ (y^\prime,Y)
	\end{equation}
	
	Where $\lambda$ is the penalty coefficient, and its value ranges from $(0,1)$. When $\lambda$ is 1, Equation \ref{loss2} is equivalent to equation \ref{loss1}.
	
	In the unlearning process, we want to minimize the loss function so that the $P(\omega,x)$ iterate to $P(\theta,x)$, thus making the $M_{initial}$ iterate in the direction of $M_0$, eliminating the training trace of $D_f$ data.

	\begin{equation}
	\label{min_loss}
		min\lambda\ L_{KL}\ (P(\omega,x),P(\theta,x))+(1-\lambda)H_{CE}\ (y\prime,Y)	
	\end{equation}
	
	\begin{equation}
	\label{st}
		s.t.P(\omega,x)={y_i\ |y_i\gets f_{\theta_t}(x_i),x\in D_f}
	\end{equation}
	
	\begin{equation}
	\label{get}
		y\prime\gets f_\theta\ (x\prime)
	\end{equation}

	\begin{algorithm} [!t]
		\caption{unlearning process}
		\label{alg:unlearning}
		\begin{algorithmic}[1]
			\footnotesize
			\STATE{ Input: initial model $f_{\theta}$ and its weight $\theta$, model for reference $h_{\theta_{0}}$ and its weight $\theta_{0}$, forget samples $D_f$, remain samples $D_r$, subset of remain samples $D_s$, penalty coefficient $\lambda$}
			\STATE{ Output:  final model $f_{\theta_{T}}$}
			
			\FOR{$t \gets 1 \text{ to T}$}  
			\STATE{ ${P(\omega,x)} \gets \text{$h_{\theta_{0}}(x)$}$ }
			\STATE{ ${P(\theta_{t-1},x)} \gets \text{$f_{\theta_{t-1}}(X)$}$}   
			\STATE{ ${d_t} \gets \text{$\lambda {L_{KL}}({P(\omega,x)},{P(\theta_{t-1},x)}) + (1-\lambda) H_{CE}(y',Y)$}$}
			\STATE{ ${\theta_t} \gets \text{$d_t(\theta_{t-1})$}$}
			\ENDFOR{ delete $D_f$ from initial model}
			\RETURN{ $f_{\theta_{T}}$}
		\end{algorithmic}
	\end{algorithm}			
	
	As shown in Algorithm \ref{alg:unlearning}, firstly, we input the sample $x$ of each forgotten dataset to the reference model $M_0$ and calculate the output distribution $P(\omega,x)$. Meanwhile, we input $x$ to the current model $f_{\theta_t}$ (updated after iteration $t-1$) and calculate the output distribution $P(\theta_t,x)$. At iteration $t$, equation \ref{loss2} is used to obtain the parameter after the $t$ time update ($H_{CE}$ comes from equation \ref{ce}), and the forgotten final model $f_{\theta_T}$ is obtained after $T$ times of iteration.

	\subsection{REFERENCE MODEL}
	
	In our approach, we view unlearning as a process of moving towards a model without  data to be forgotten training. It is important to have a reference model $M_0$, that provides the direction of change for the target model. The reference model should not use the forgotten data $D_f$ as a training set to ensure that it provides directions that lead the target model to eventually forget $D_f$. For $M_0$, $D_f$ does not belong to its training set, so when $D_f$ input is calculated, the output distribution $P(\omega,x)$ obtained is not similar to that obtained by training data. When the target model of the output of the $D_f$ distribution $P(\theta,x)$ is nearly equal to the reference model for the output of the $D_f$ distribution, $D_f$ for target model, the output distribution $P(\theta,x)$ has not been similar with other output distribution of training data, we think $D_f$ is no longer have as members of the training of the model, the influence of this part of the data at this time has been forgotten. If $D_f$ cannot only be the training data as the condition of $M_0$, then selecting a model whose weight distribution is too far from that of the target model for training can achieve the goal of unlearning, but it will have a serious impact on the weight of the target model, destroy the performance of the target model itself and make it unable to work normally.
	Therefore, when selecting the reference model $M_0$, it is also necessary to consider that the weight distribution of the reference model should be as similar as possible to that of the target model, so that the adjustment of the forgotten target model will not affect its own performance in a small range. Ideally, a model that deletes $D_f$ of data to be forgotten and uses $D_r$ for training with all remaining data has the same training data as the target model, so the weight distribution is also the same, and the direction provided is the natural optimal choice. However, it takes a lot of time and space to retrain the model, so we hope to find a reference model $M_0$ with the same effect and as little cost as possible. Therefore, in our method, $D_s$, a subset of the remaining data $D_r$, is adopted as the training set of the reference model, and the time cost of training the reference model is reduced under the condition that the distribution difference with the re-training model is small.\\
	DEFINITION 2 REFERENCE MODEL:
	
	Given a dataset $D$, we need to forget dataset $D_f$, leaving $D_r$,$D_r\cup D_f=D$.The parameter of the model $h$ trained by $D_r$ is $\varepsilon$.We assume that (model $h$ obeys a posterior distribution $P$)  $h_{\varepsilon} (D_r)\sim{Q}$. If a model $r$ satisfies $r(D_r)\sim{Q\prime}$ and $KL(Q,Q\prime)\approx0$, then $r$ is considered to be the reference model.

	\subsection{EVALUATION}

	\begin{algorithm} [!t]
		\caption{backdoor process}
		\label{alg:backdoor}
		\begin{algorithmic}[1]
			\footnotesize
			\STATE{ Input: initial model $f_{\theta}$ and its weight $\theta$, backdoor target class $y_0$, forget samples $D_f$, remain samples $D_r$, remain samples $D_r$}
				\STATE{ Output:  final model $f_{\theta}$}
				\STATE{$\forall x,y \in{D_f}$}
				\WHILE{x==True} 
				\STATE{$x={x\prime}$}
				\STATE{$y={y_0}$}
				\ENDWHILE
				\STATE{train $f_{\theta}$ with $D_f$ and $D_r$}
				\IF{$f_{\theta}(x\prime)=={y_0}$}
				\STATE{attack success}
				\ELSE
				\STATE{attack fail}
				\ENDIF
				\RETURN{ $f_{\theta}$}
		\end{algorithmic}
	\end{algorithm}

	\begin{figure*}[!h]
		\centering
		\includegraphics[width=0.8\textwidth]{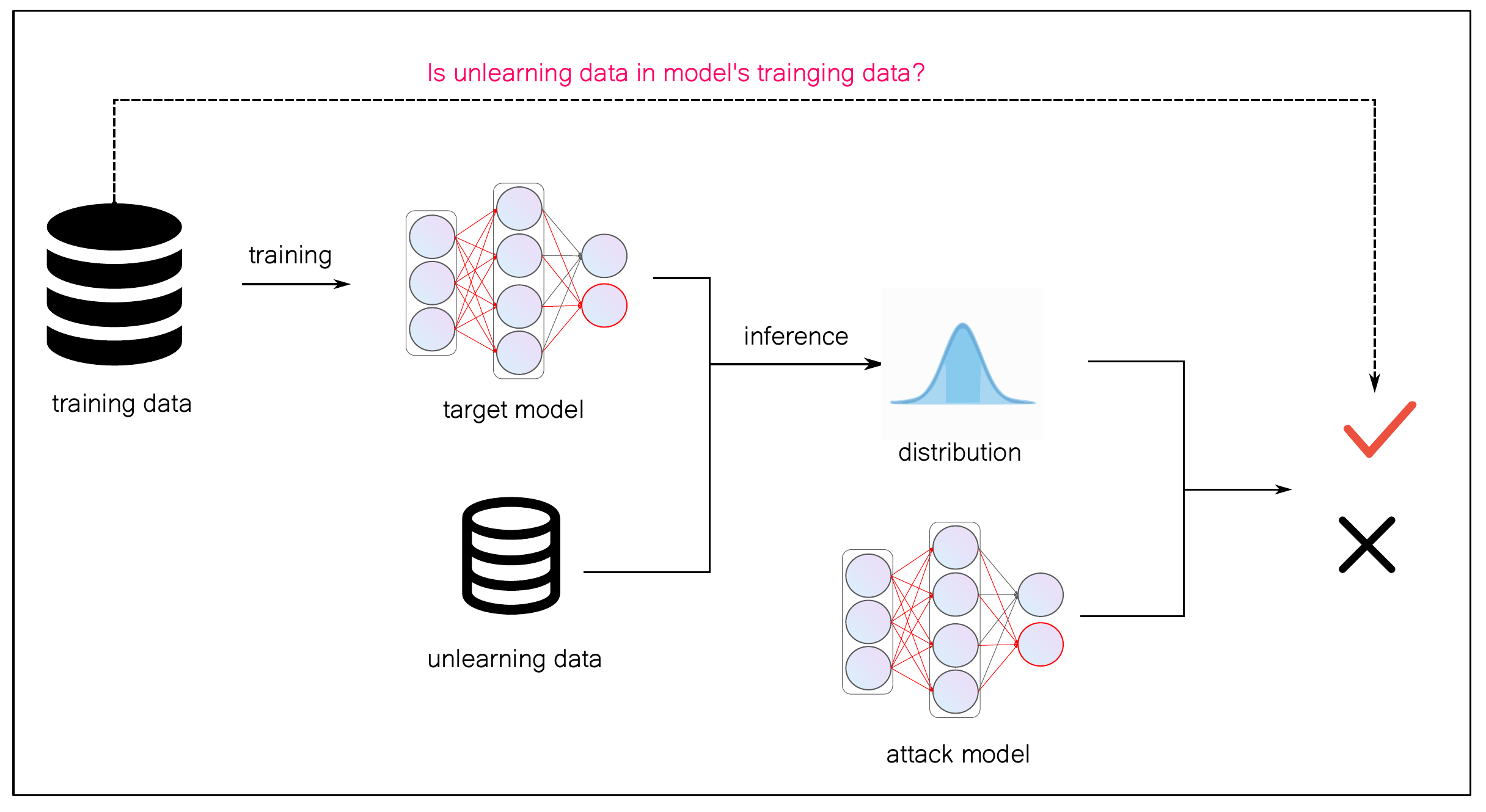}
		\caption{membership attack}
		\label{fig:membership}
	\end{figure*}

	After unlearning, one more task is required to determine whether the impact of unlearning data on the model has been completely eliminated. Membership inference attack can judge whether a sample exists in a model's training set according to its distribution. Membership inference attack is realized based on a shadow model. Firstly, a shadow dataset similar to the target model's training set distribution needs to be integrated into the shadow dataset. With this shadow dataset as the training set, a shadow model can be trained. At this point, continue to prepare the shadow model, shadow training set, shadow non-training set.
	
	On this basis, a binary attack model is trained. Given an input, the attack model can judge whether it is a member of the training set of the target model. In our method, member inference attack \cite{2017Membership} is used to evaluate the residual unlearning information in the model after unlearning.In\cite{2017Membership}, the attack model of member reasoning is trained based on the original model before unlearning, which can accurately distinguish learned or unlearned data and judge whether it is member data according to the posterior distribution difference between the input sample and the remaining data $D_r$. In our experiment, as shown in the figure \ref{fig:membership}, input the data to be forgotten into the final model, calculate the output distribution, and the attack model determines whether it is the member data of the target model according to this distribution. If the attack model deduces that the forgotten data is non-member data, the forgotten data is successfully forgotten.\\
	DEFINITION 3 MEMBERSHIP INFERENCE ATTACK:
	
	Suppose the member inference attack algorithm is $\partial$, given a target model $f$, when $\partial(f(x))=True$, x is a member of target model's training data, when $\partial(f(x))=False$, it is not.
	
\begin{figure*}[ht]
		\centering
		\includegraphics[width=0.5\textwidth]{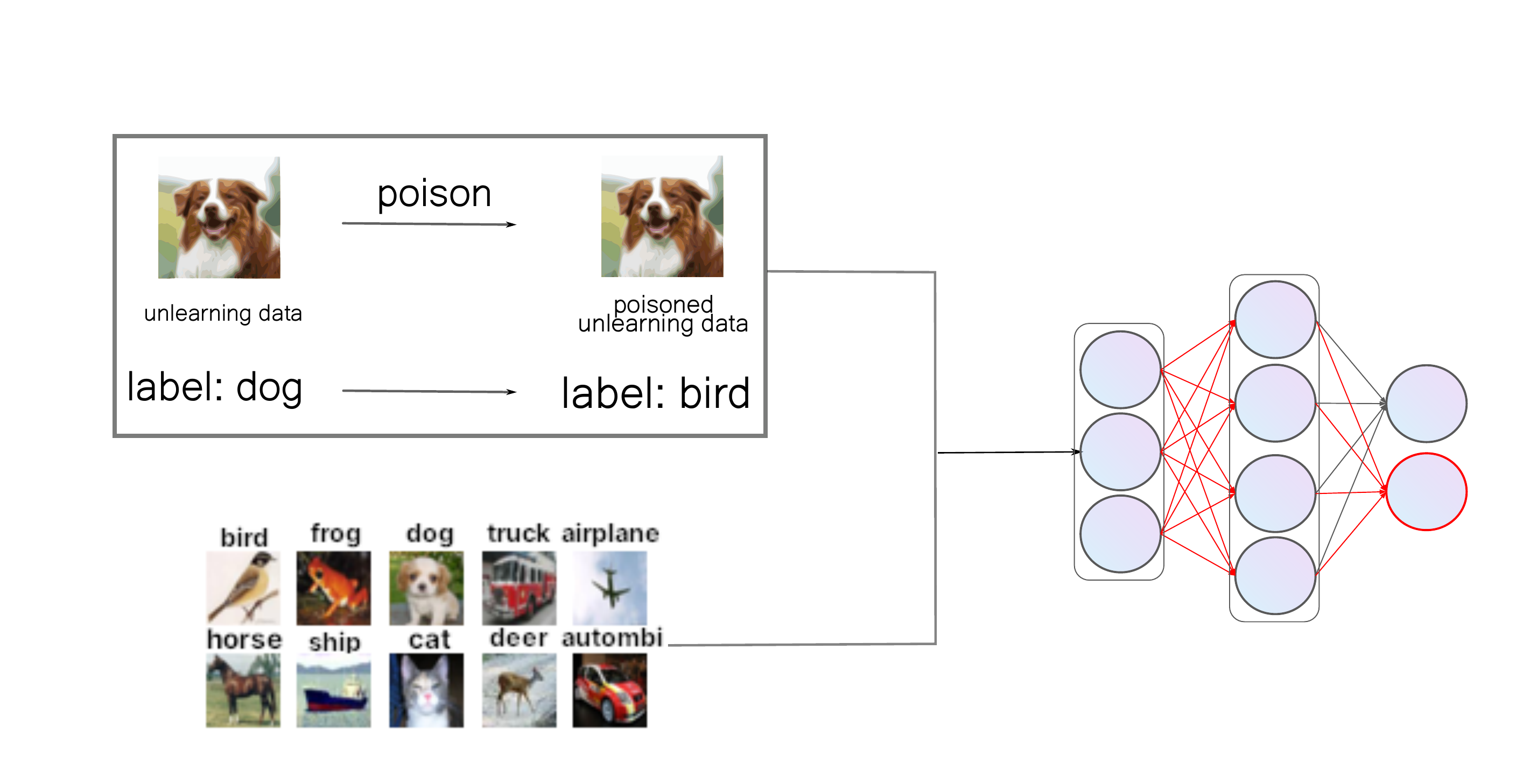}
		\caption{implant the trigger into target model before training}
		\label{fig:backdoor1}
	\end{figure*}

\begin{figure*}[ht]
		\centering
		\includegraphics[width=0.5\textwidth]{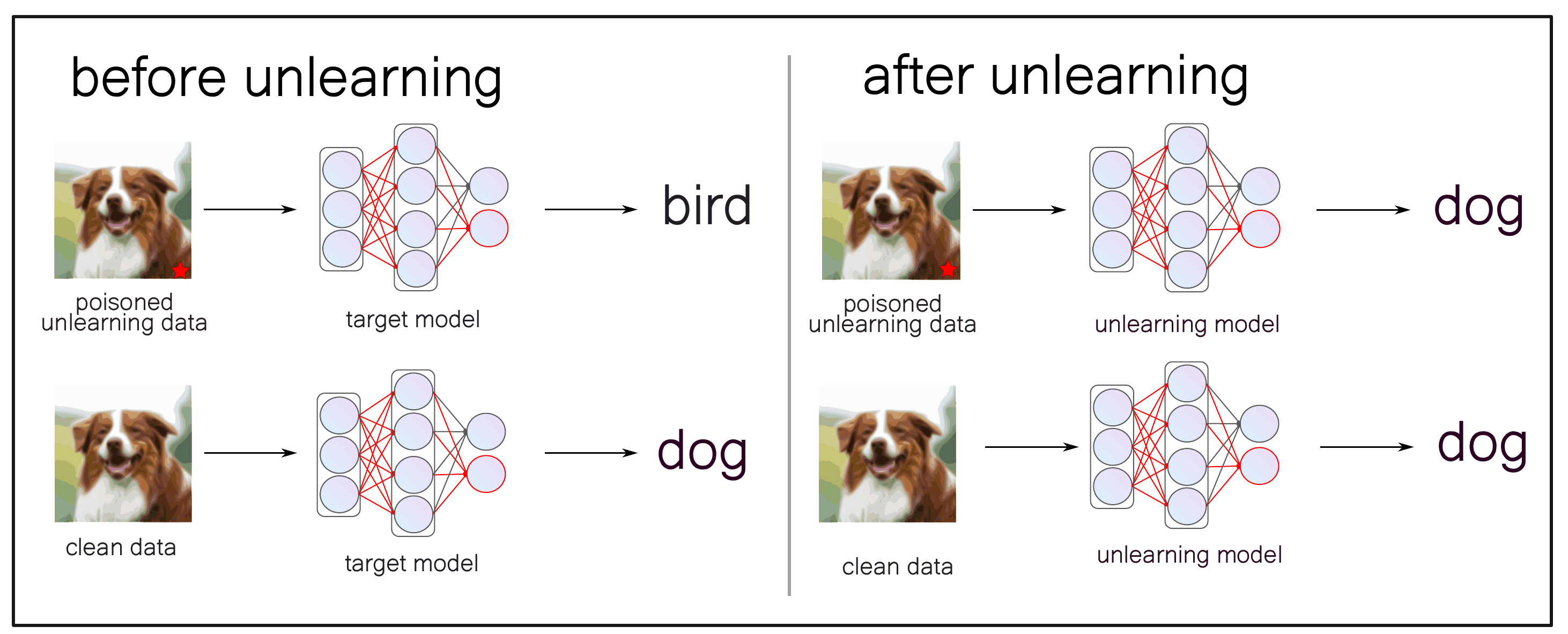}
		\caption{backdoor attack}
		\label{fig:backdoor2}
	\end{figure*}

	As a common attack method, backdoor attack is helpful to evaluate the effect of unlearning. \cite{2017BadNets} insert a backdoor into part of the data, and the data implanted by the backdoor will be uniformly labeled with a specific label. The model generated by backdoor data training "remembers" the backdoor information, which will be triggered when encountering the same backdoor and produce the same prediction result, i.e. fixed label.Apply this principle in our experiment. As figure \ref{fig:backdoor1}, first, the backdoor was implanted into the forgotten data $D_f$, and the label was set as a specific label, which was trained to generate the original model $M$ together with the normal data (namely the residual data $D_r$).
	As shown in figure \ref{fig:backdoor2}, on the left, $M$ at this time can predict normally when it encounters normal test data, and will generate the specified label when it encounters test data with the same backdoor. It is applied in the unlearning evaluation stage, assuming that we already have $M$ trained by backdoor unlearning data and normal residual data, then we use our method to forget, ideally, as shown in the figure \ref{fig:backdoor2} on the right, the forgotten model $M_{final}$ will no longer be affected by the forgotten data $D_f$, that is, the data with the backdoor has been forgotten. At this point, after the test data with the back door is input into $M_{final}$, the probability distribution obtained will be similar to that of normal data, and the same prediction result will be generated as that of the rear door of the device, instead of the specific label on the back door. Therefore, the test accuracy of this part of the test data with backdoor will be low, keeping around one of the total number of categories. Conversely, when the data is not completely cleared, the model still "remembers" the backdoor information, and when it encounters test data, its probability distribution will differ greatly from that of normal data, and it will often predict the specific label stamped on the backdoor. Therefore, the test accuracy of this part of test data with backdoor will be higher.

	DEFINITION 4 BACKDOOR ATTACK:
	
	Given a algorithm of backdoor attack $\Omega$, a target model $f$, its parameters $\theta$, clean training data $(x,y)\in D$, poisoned training data $(x\prime,y\prime)\in D_f$. If there is $\Omega(f_{\theta_T})=y\prime$, then attack success. Else $\Omega(f_\theta\ (x\prime))=y$, attack fail.
	In our method,input the unlearning data into the model after unlearning $f_{\theta_T}$, if there is $\Omega(f_{\theta_T}(x\prime))=y$, unlearn success. Else $\Omega(f_{\theta_T}\ (x\prime))=y\prime$, unlearning fail.

	
	\section{Experimental Results}	
	
	\subsection{Experimental Setup}
	
	\begin{enumerate}
		\item \textbf{DATASET:}In the experiment, MNIST, fashion-MnIST, Cifar10 and SVHN datasets were adopted.The size of the forgotten data samples was 1/100 of the number of samples in the dataset, which were 600,600,500,732, respectively.

		\item\textbf{MODEL:}We tested the effect of our method on different models, including multilayer perceptron, Lenet, ResNET-18,VGG16,wide resnet, as shown in the table below \ref{fig:model}.

		\renewcommand\arraystretch{1.5}
		\begin{table}[!h] 
			\centering
			\caption{Datasets and models used in the experiment}
			\label{fig:model}  
			\scalebox{0.6}{
				\begin{tabular}{p{3.2cm}cccc} 
					\hline 
					dataset & model architecture & number of instance &classes\\
					\hline
					Mnist		  & Lenet 		 & 60000	 & 10\\
					Fashion-mnist & MLP 		 & 60000	 & 10\\
					SVHN		  & wide resnet 	 & 73257	 & 10\\
					Cifar10.R	  & Resnet18 	 & 50000	 & 10\\
					Cifar10.V	  & VGG16 		 & 50000 	 & 10\\
					\hline
			\end{tabular}}  
		\end{table}

		Mnist: MNIST dataset is the benchmark dataset for segmentation and center handwritten digital gray image \cite{1998Gradient}. We used 60,000 training examples, and 10,000 test examples, with each image being 28×28 pixels in size. In the selection of target model, we use Lenet-5 \cite{1998Gradient}, which is a network architecture composed of 2* CONV, 1*pool and 1*FC, and a very efficient convolutional neural network for handwritten character recognition.
		
		Fashion-MNIST is a data set of Zalando article images \cite{2017Fashion}, including a training set of 60,000 examples and a test set of 10,000 examples. Each example is a $28\times28$ gray scale image associated with labels from 10 categories, each of which is a clothing item. The dataset shares the same image size and training and test segmentation structure with MNIST. In the selection of target model, we use multi-layer perceptron.

		Cifar-10 is a base quasi-data for evaluating image recognition algorithms. The dataset consists of 60,000 color images of three channels with a size of 32×32, divided into 10 categories such as "aircraft", "dog" and "cat".  In particular, CIFAR-10 is a balanced dataset with 6000 randomly selected images per category. In the CIFAR-10 dataset, we used 50000 training images and 10000 test images. In the selection of cifar-10 target model, Resnet18 was used \cite{2016Deep}. In order to better compare the influences brought by different models, we also used VGG16 \cite{2014Very} as the target model in CIFAR-10 dataset.
		
		SVHN (StreetViewHouseNumber) dataset \cite{netzer2011reading} comes from Google StreetView number. The training set consisted of 73,257 32×32 three-channel color images with the numbers 1 to 10. The test set consisted of 26,032 images. In the selection of target model of SVHN dataset, we adopt multilayer perceptron.

	\end{enumerate}
	
	\subsection{Evaluation indicator}
	When $\lambda$ and the reference model are selected (see table \ref{fig:default}), we evaluate the performance of the forgotten model in our method in terms of the accuracy after unlearning, member inference attack and backdoor attack, and the time required for unlearning.
	
	\renewcommand\arraystretch{1.5}
	\begin{table}[!h] 
		\centering
		\caption{Default parameter settings}
		\scalebox{0.6}{
			\begin{tabular}{p{3.2cm}ccccc} 
				\hline 
				dataset & model architecture & number of instance & value of $\lambda$ &numer of instance in M\\
				\hline
				Mnist		  & Lenet 		 & 60000	 & 0.01		&6000\\
				Fashion-mnist & MLP 		 & 60000	 & 0.01		&6000\\
				SVHN		  & wide resnet 	 & 73257	 & 0.01		&7325\\
				Cifar10.R	  & Resnet18 	 & 50000	 & 0.01		&5000\\
				Cifar10.V	  & VGG16 		 & 50000 	 & 0.01		&5000\\
				\hline
		\end{tabular}}
		\label{fig:default}    
	\end{table}

	\begin{enumerate}
		\item \textbf{ACCURACY:}Accuracy is an important indicator of target model training. In order to apply machine unlearning into practice without affecting the actual effect of target model, it is necessary to evaluate the accuracy before and after unlearning. Accuracy was defined as accuracy= (TP+TN) /(P+N), the number of correctly classified samples divided by the number of all samples.1) True positives(TP): the number of Truepositives that were correctly classified as positive cases, that is, the number of examples that were actually positive cases and were classified as positive cases by the classifier;2) False positives: the number of instances incorrectly classified as positive by a classifier;3) False negatives(FN): the number of instances incorrectly classified as negative, i.e. the number of instances that are actually positive but classified as negative by the classifier;4) True negatives(TN): the number of instances correctly classified as negative, i.e. the number of instances that are actually negative and classified as negative by the classifier.
		
		The accuracy rate of different datasets after unlearning is shown in figure \ref{fig:accuracy-comparison}. As shown, for MNIST dataset, the accuracy of the first column is 98.97\% before unlearning, and the accuracy of the second column is 98.96\% after retraining after deleting 1/100 of the samples. Since the sample size is not greatly reduced, the accuracy of the third column is 98.12\% after unlearning through our method. Compared with the retrained model, the accuracy is not significantly decreased, indicating that our method does not affect the performance of the target model itself. For the performance of CIFAR10 dataset on ResNet, the accuracy of the first column before unlearning is 90.87\%, and the accuracy of the second column after retraining is 4\% lower than that before unlearning due to the reduction of sample number. In our method, the accuracy of the model after unlearning is 89.64\%, slightly higher than that of the retrained model. For the performance of CIFAR10 dataset on VGG16, the trend before and after unlearning is the same as that on RESNET18, but the overall accuracy is less than 3\% than that of RESNET. This is because ciFAR10 dataset of VGG16 model training has its own architectural limitations compared with RESNET18 model and is not affected by the unlearning method.
		
		In the figure \ref{fig:accuracy-comparison}, we can see that for svhn, cifar. R and cifar. V datasets, the accuracy of our method after unlearning will be higher than that of retraining, which is caused by uneven sampling. As shown in the figure \ref{fig:sampling}, all points in the figure represent the whole data set before unlearning, the blue dot represents the correctly classified samples, a total of 80 correct samples, the yellow dot represents the incorrectly classified samples, a total of 20 correct samples, and the correct classification green is 80\%. The data to be forgotten obtained by random sampling may be distributed in the circle, with a total of 9 correct samples and 1 wrong sample, and the classification accuracy is 90\%. This part of the data to be forgotten is removed, and the remaining data are retrained. There are 71 correct samples and 19 wrong samples, and the classification accuracy is 78.8\%. It can be seen that random sampling may lead to higher accuracy after unlearning than retraining, or even higher than that before unlearning.

		\begin{figure}[!h]
			\centering
			\includegraphics[width=0.5\textwidth]{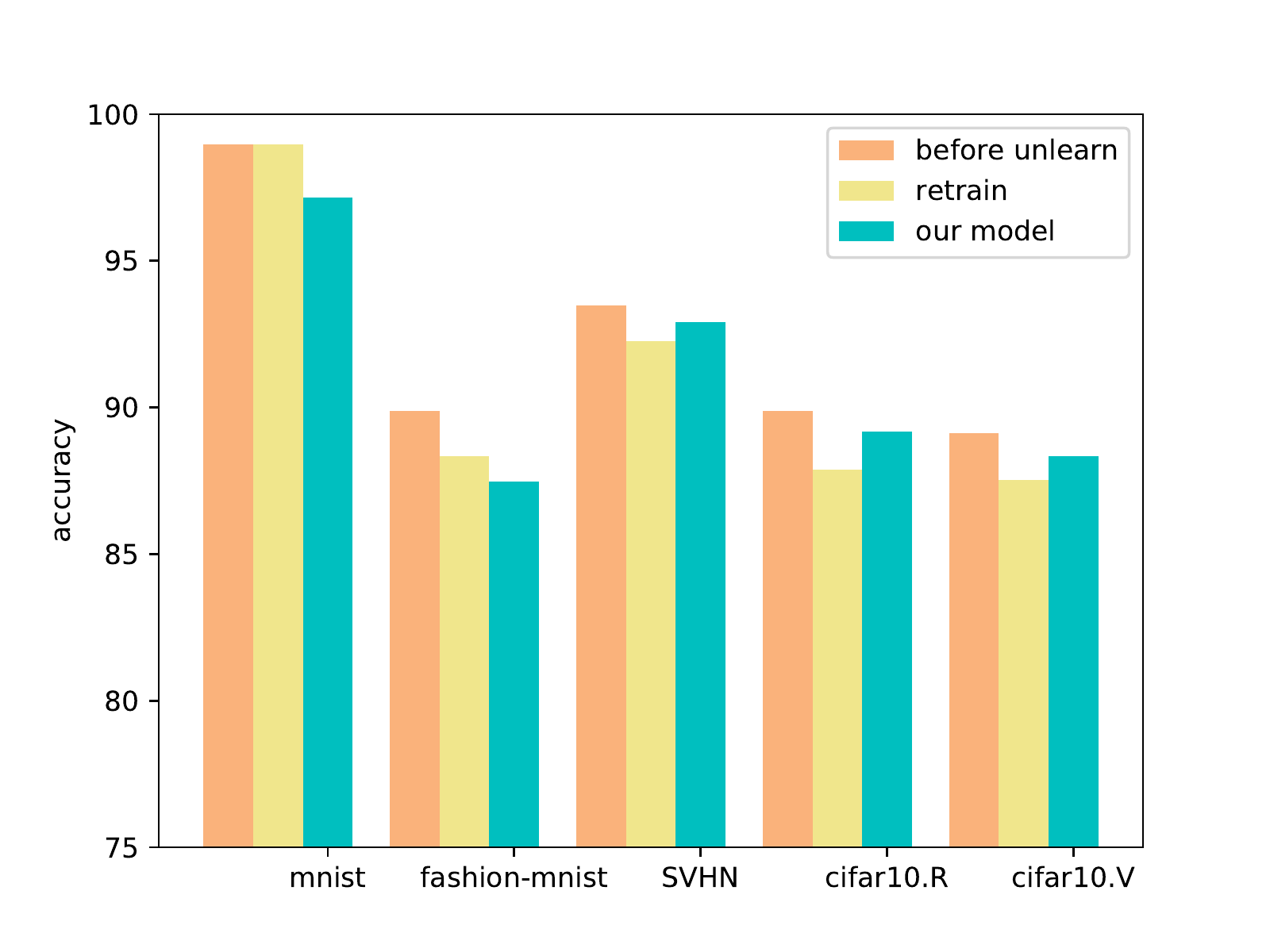}
			\caption{Comparison of accuracy of different datasets before and after unlearning}
			\label{fig:accuracy-comparison}
		\end{figure}
	
		\begin{figure}[!h]
		\centering
		\includegraphics[width=0.5\textwidth]{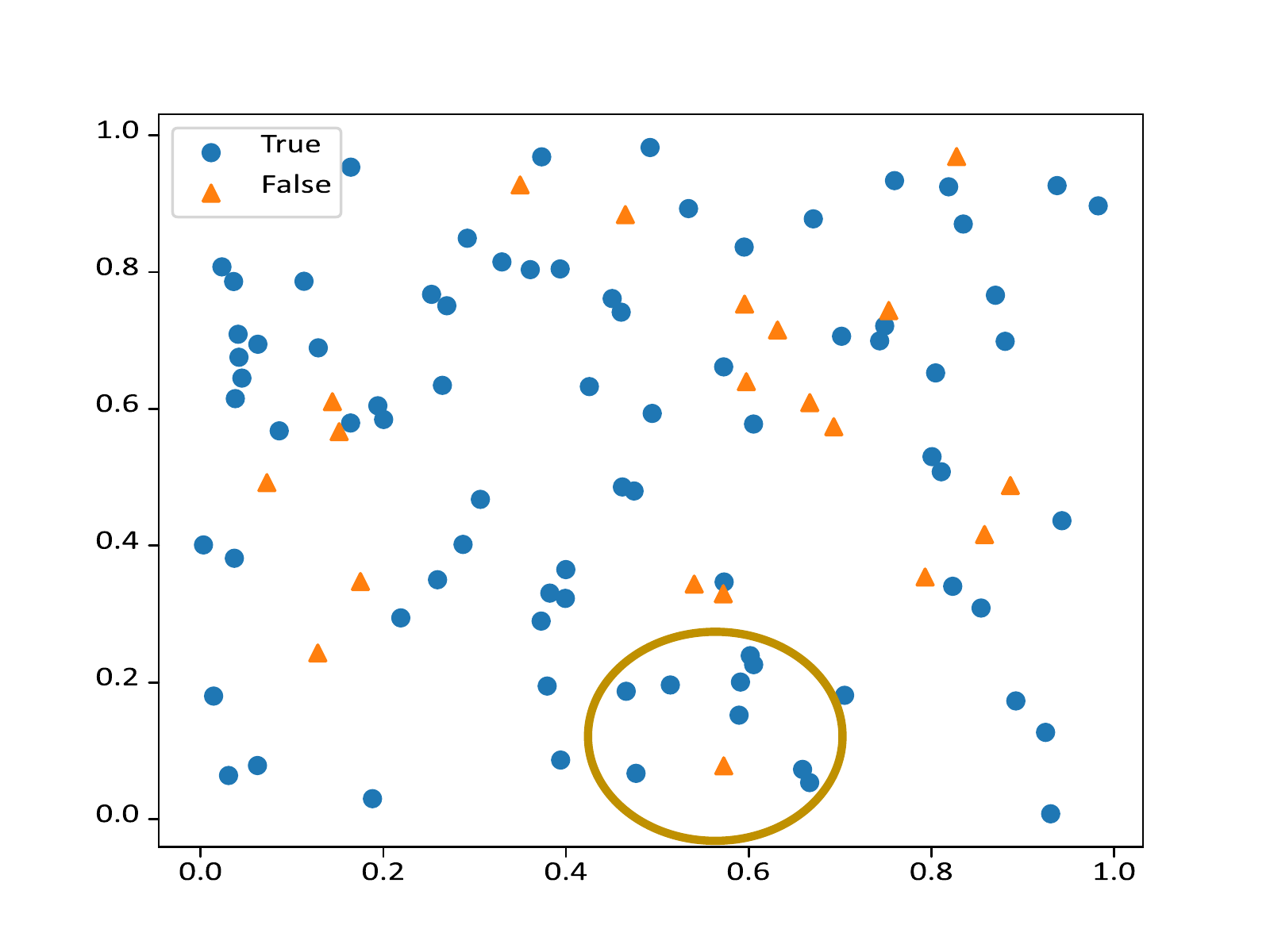}
		\caption{Data sampling}
		\label{fig:sampling}
		\end{figure}
		
		\item \textbf{MEMBERSHIP ATTACK:}
		In the inference stage, input the data to be forgotten into the attack model, and the attack model deduces whether it is the member number data of the target model according to the distribution of the output. If the accuracy of member inference is too high or too low, it means that the attack model has a higher confidence to identify forgotten data as member or non-member data. Before unlearning, the data to be forgotten is a part of the target model member data, so the posterior distribution of this part of the data is naturally similar to other member data, and can be easily identified as member data for the attack model, as shown in figure \ref{fig:membership-comparison}. As shown, its accuracy can reach 90\%, which means that the proportion of the attack model correctly identifying the data to be forgotten as member data is 90\%. In the retrained model, the data to be forgotten by is no longer the member data of the model. As the posterior distribution of differs significantly from that of member data, it is difficult for the attack model to judge whether it is a member data or not, as shown in the figure, the accuracy rate of is about 50\%. After using our method for unlearning , the accuracy of the attack did not differ significantly from that of the retraining method. This means that for the attack model, the data to be forgotten is difficult to be clearly identified as member or non-member data. In other words, the part of data to be forgotten at this time "does not remember" the information of the target model for the forgotten model, just as for the retrained model.

		\begin{figure}[!h]
			\centering
			\includegraphics[width=0.5\textwidth]{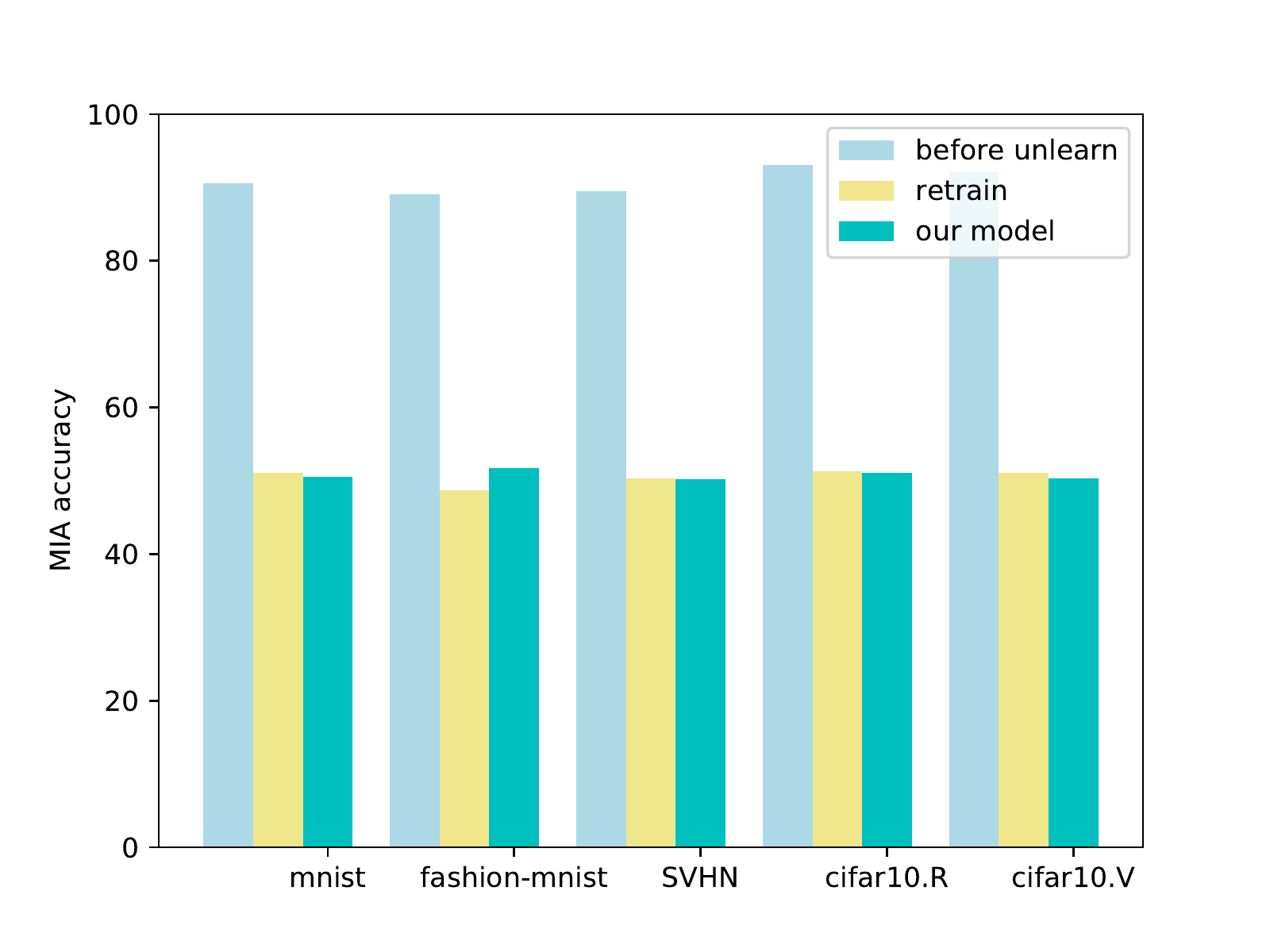}
			\caption{Comparison of membership attack accuracy after unlearning different datasets}
			\label{fig:membership-comparison}
		\end{figure}
		
		\item \textbf{BACKDOOR ATTACK:}
		To determine whether unlearning is complete, we use a backdoor attack to evaluate the effect of unlearning. In the experiment, we first implanted the data to be forgotten into the back door, labeled with fixed labels, and used the remaining normal data as the training data of the target model. At this point, the target model completed training has "remembered" the forgotten data with the back door, so the reasoning results of the model for 98\% of the back door data (that is, the data to be forgotten) are consistent with the fixed tag set by the back door, and the test accuracy of this part of the data is up to 98\%.The next step is to use our method to forget this target model and get the target model after unlearning.When the data with the back door is input into calculation again, the accuracy is less than 20\%, indicating that the reasoning result of the model for 80\% of the back door data (i.e. the data to be forgotten) is not consistent with the fixed label set. The test accuracy of data to be forgotten decreased from 98\% to 20\%, indicating that our method did make the target model "forget" the backdoor information "remembered" during the training phase, thus proving that unlearning succeeded. As shown in figure \ref{fig:backdoor-comparison},
		 during the unlearning process of MNIST dataset, the test accuracy of the data that do not need to be forgotten remains at a high level, which also reflects that the accuracy of the remaining data after unlearning remains at a high level. On the forgetting data, the test accuracy decreased from the initial 98 \% to a low level, representing the unlearning effect. Weigh the unlearning performance and precision after unlearning, and assign a weight of 0.5 respectively. The higher the result, the best overall performance, as shown by the blue line in figure \ref{fig:backdoor-comparison}. The same is true for the cifar10.R dataset, as shown in figure \ref{fig:backdoor-comparison2}.

		\begin{figure}[!h]
			\centering
			\includegraphics[width=0.5\textwidth]{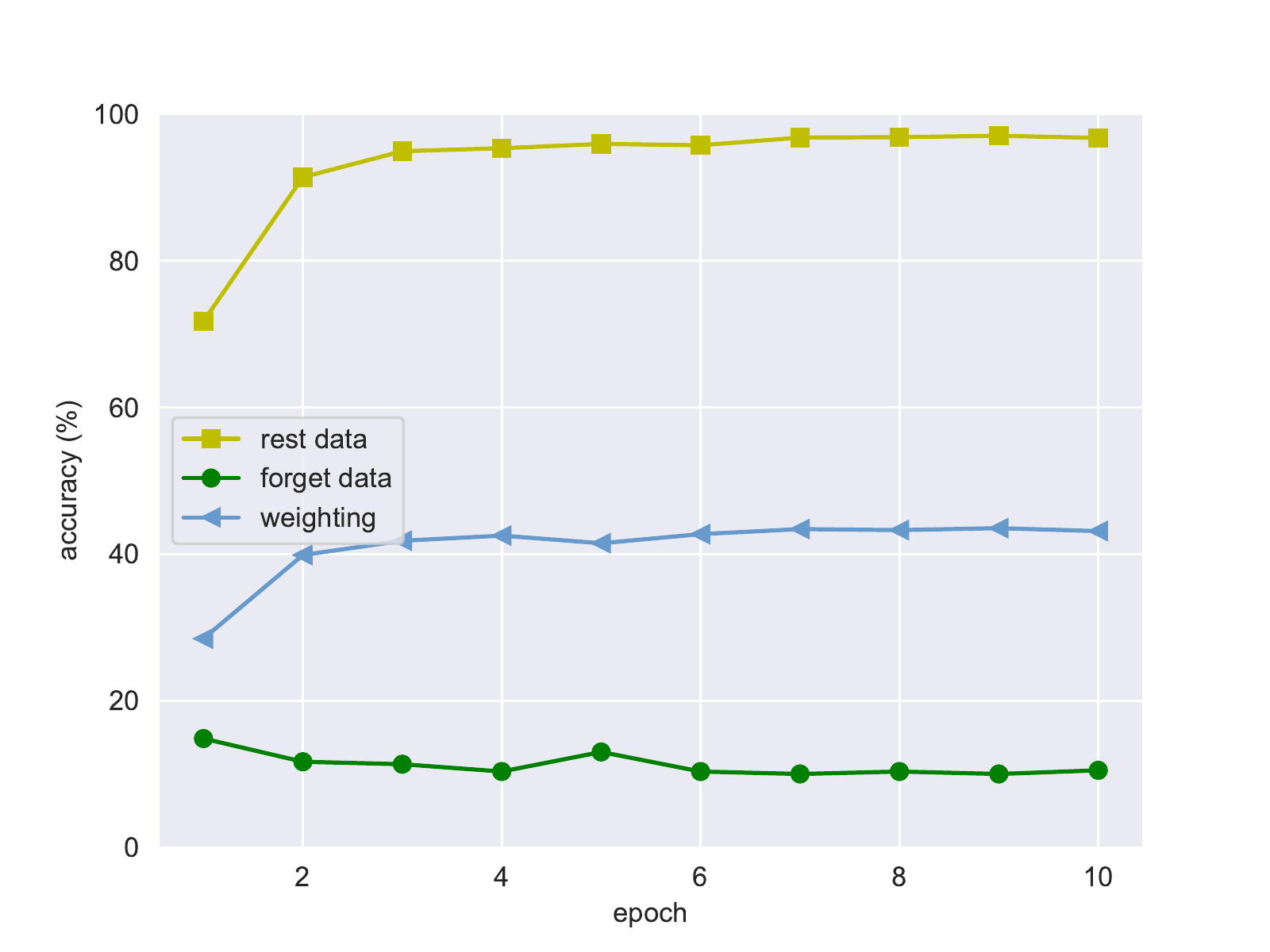}
			\caption{Comparison of backdoor attack test accuracy after mnist dataset unlearning}
			\label{fig:backdoor-comparison}
		\end{figure}
		
		\begin{figure}[!h]
			\centering
			\includegraphics[width=0.5\textwidth]{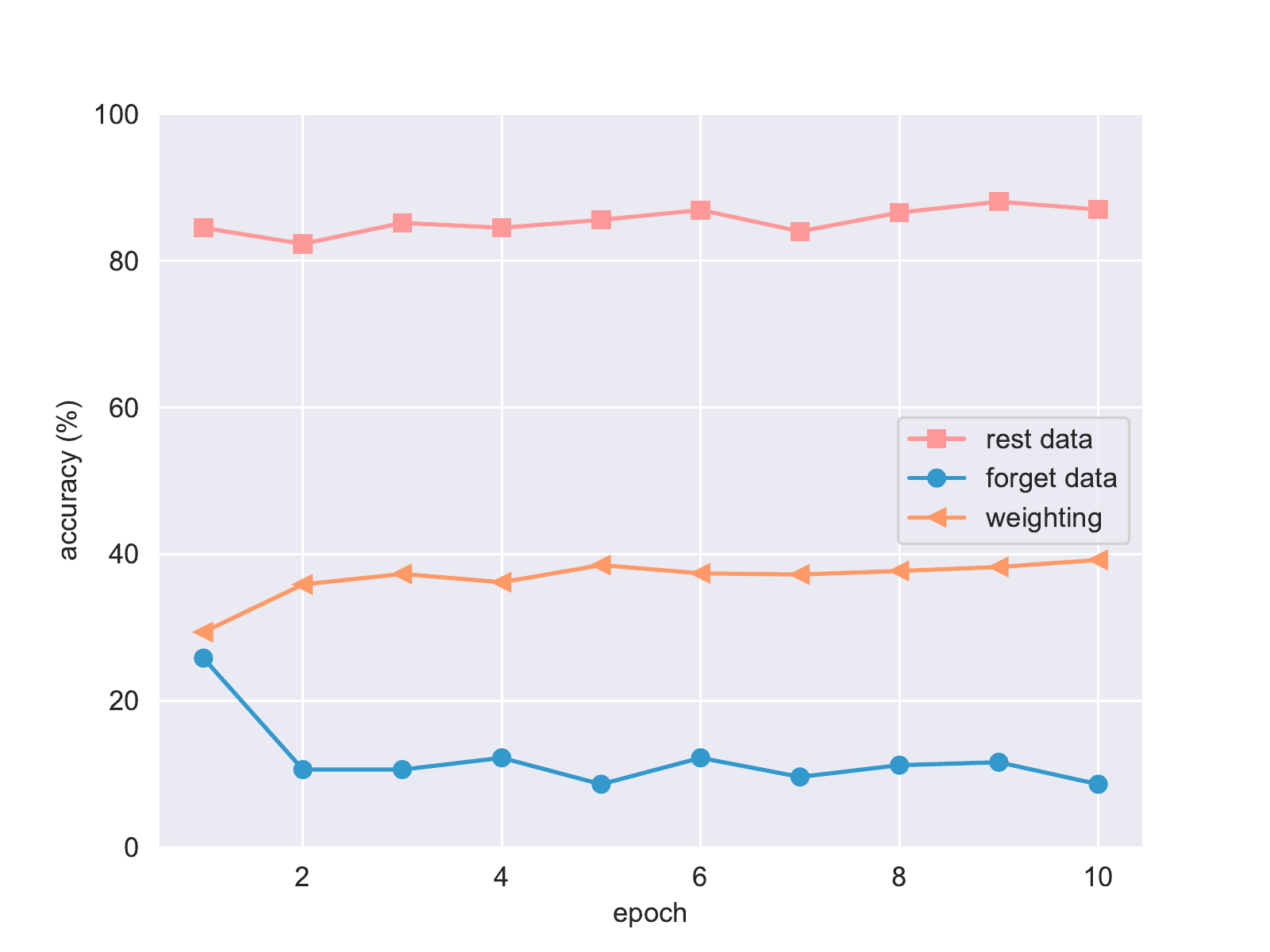}
			\caption{Comparison of backdoor attack test accuracy after cifar10.R dataset unlearning}
			\label{fig:backdoor-comparison2}
		\end{figure}
		
		\item \textbf{TIME COST:}
		In addition to performance, efficiency is a key factor in data unlearning. The goal is to achieve almost the same results as retrain in less time.  Taking the Retrain model as a reference, our method can increase the time consumption by tens of times. As table \ref{Tab:time_cost} shown, for mnist dataset, the data to be forgotten accounts for 1/100 of the total dataset, a total of 600 items. Our method requires two iterations to reach the end point, and the time required is 3.81s. It takes 42.81s to train the reference model, and the total time is 46.62s. The total time of retraining after deleting the information to be forgotten was 750.69s. By contrast, our method accelerates 16.10 times. And the same goes for everything else. 
		
		\begin{figure}[!h]
			\centering
			\includegraphics[width=0.5\textwidth]{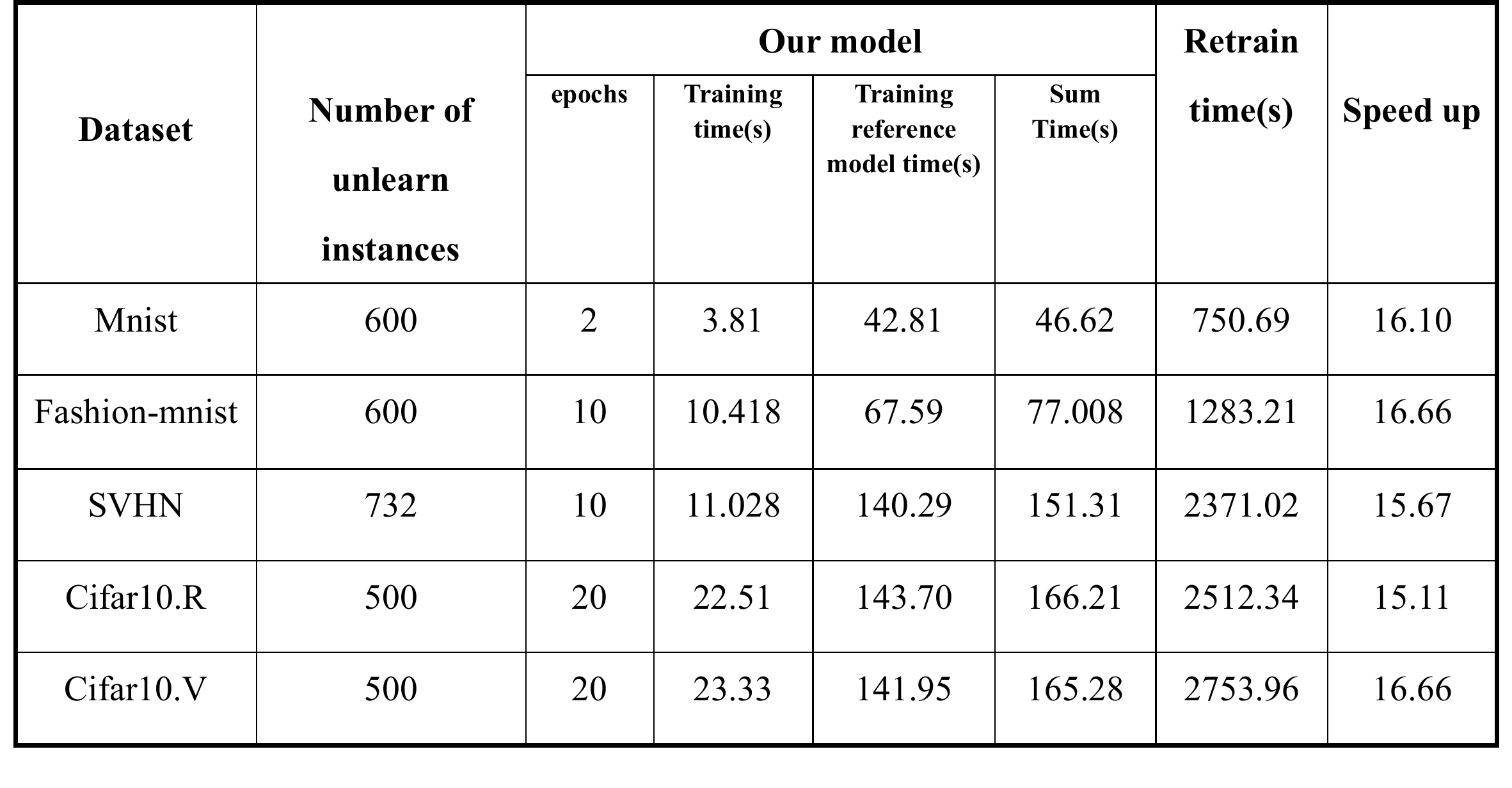}
			\caption{Comparison of unlearning time of different datasets}
			\label{Tab:time_cost}
		\end{figure}
	\end{enumerate}

	\subsection{The influence of $\lambda$}
	
\begin{figure*}[ht]
		\centering
		\includegraphics[width=7cm]{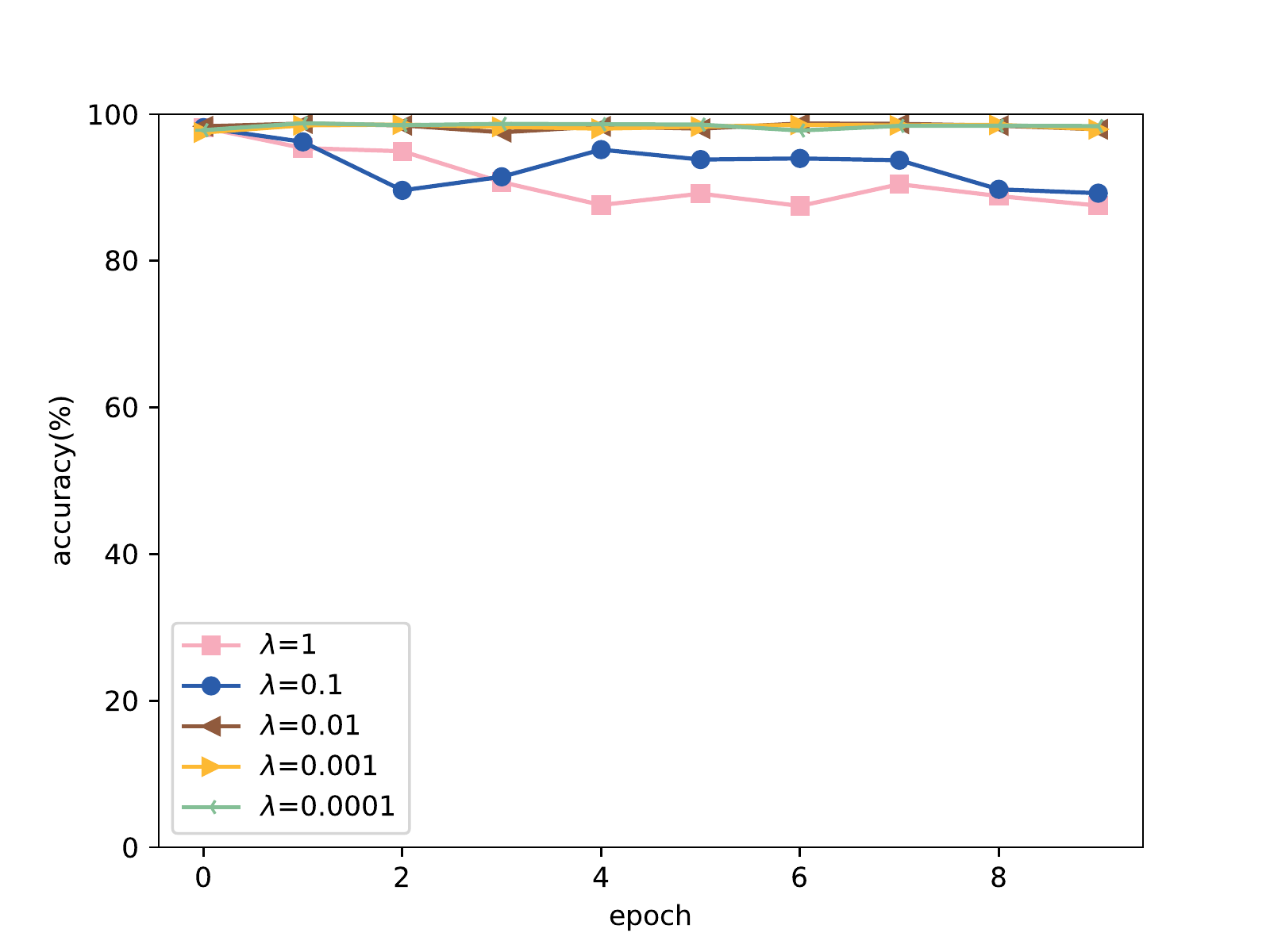}
		\hspace{0.5in}
		\includegraphics[width=7cm]{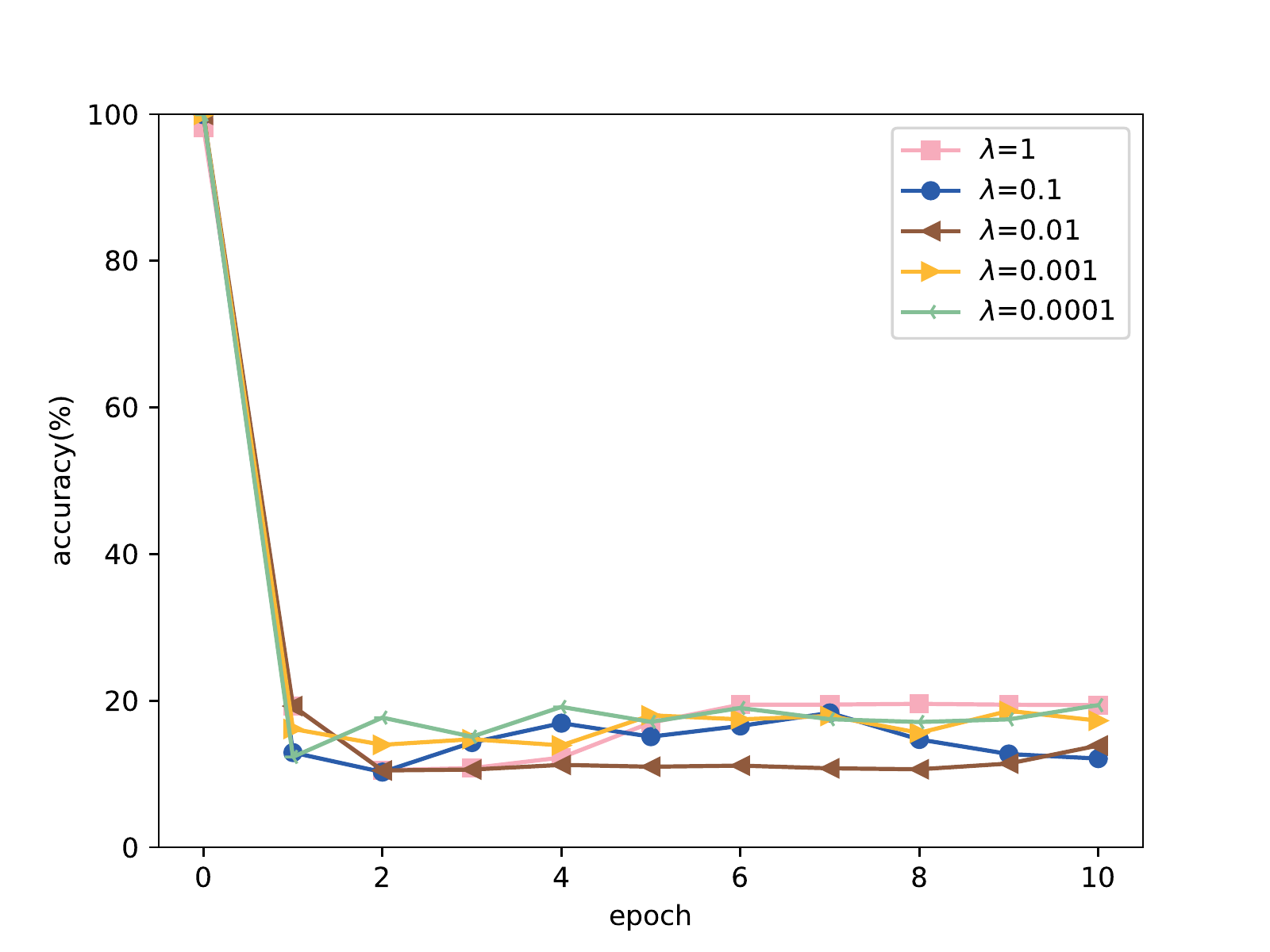}
		\caption{Influence of different of $\lambda$ values on the accuracy of MNIST dataset after unlearning (left)
			Influence of different values of $\lambda$ on the accuracy of backdoor attack after MNIST dataset unlearning (right)}
		\label{fig:minipage2}
	\end{figure*}

	As mentioned in the previous chapter, we use the precision penalty term to constrain the precision loss of the target model after unlearning (equation  \ref{loss2} to control sequence), where $\lambda$ is the coefficient of the normal unlearning term, and $1-\lambda$ is the penalty term coefficient. In order to obtain the best unlearning effect without losing too much training accuracy, it is necessary to balance the weight of the two terms. If $\lambda$ is too large, the loss of precision will be serious, while if $\lambda$ is too small, the effect of unlearning will be weakened.In the experiment, we tried the effect of different $\lambda$ on unlearning performance. As shown in figure \ref{fig:minipage2} on the left, taking mnist dataset as an example, when $\lambda$ is 1 and 0.1, the accuracy of the target model in unlearning decreases obviously. When the value is 0.01, 0.001, or 0.0001, the accuracy remains stable. As shown in figure \ref{fig:minipage2} on the right, when $\lambda$ is 0.01, the accuracy is the lowest, indicating that the unlearning effect is the best. To balance accuracy and unlearning effect, 0.01 $\lambda$ is the most suitable. For other datasets, the values were 0.01 (fashion-MNIST) and 0.001 (CIFAR), respectively.

	\begin{figure*}[ht]
		\centering
		\includegraphics[width=7cm]{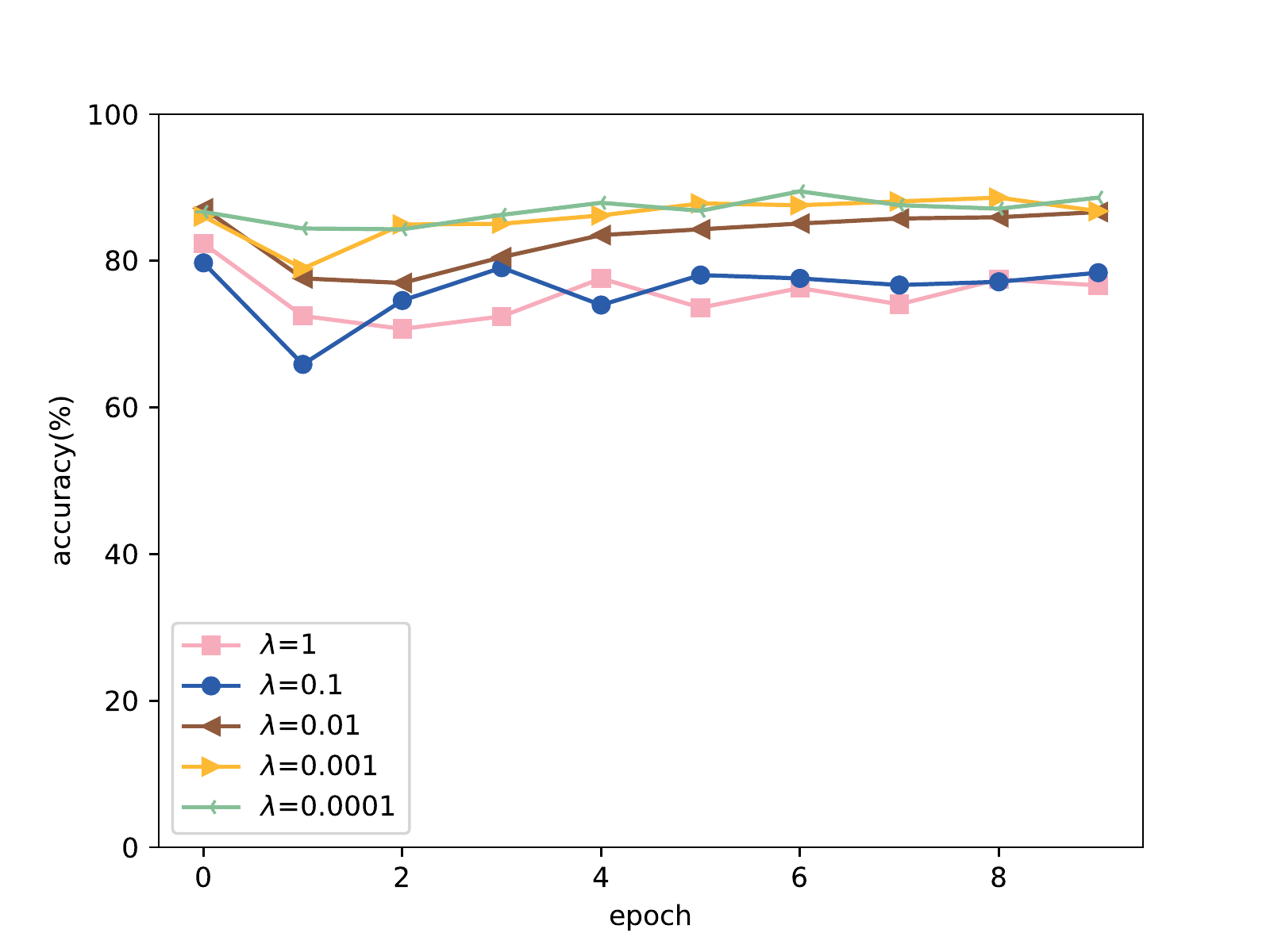}
		\label{fig:minipage3}
		\hspace{0.5in}
		\includegraphics[width=7cm]{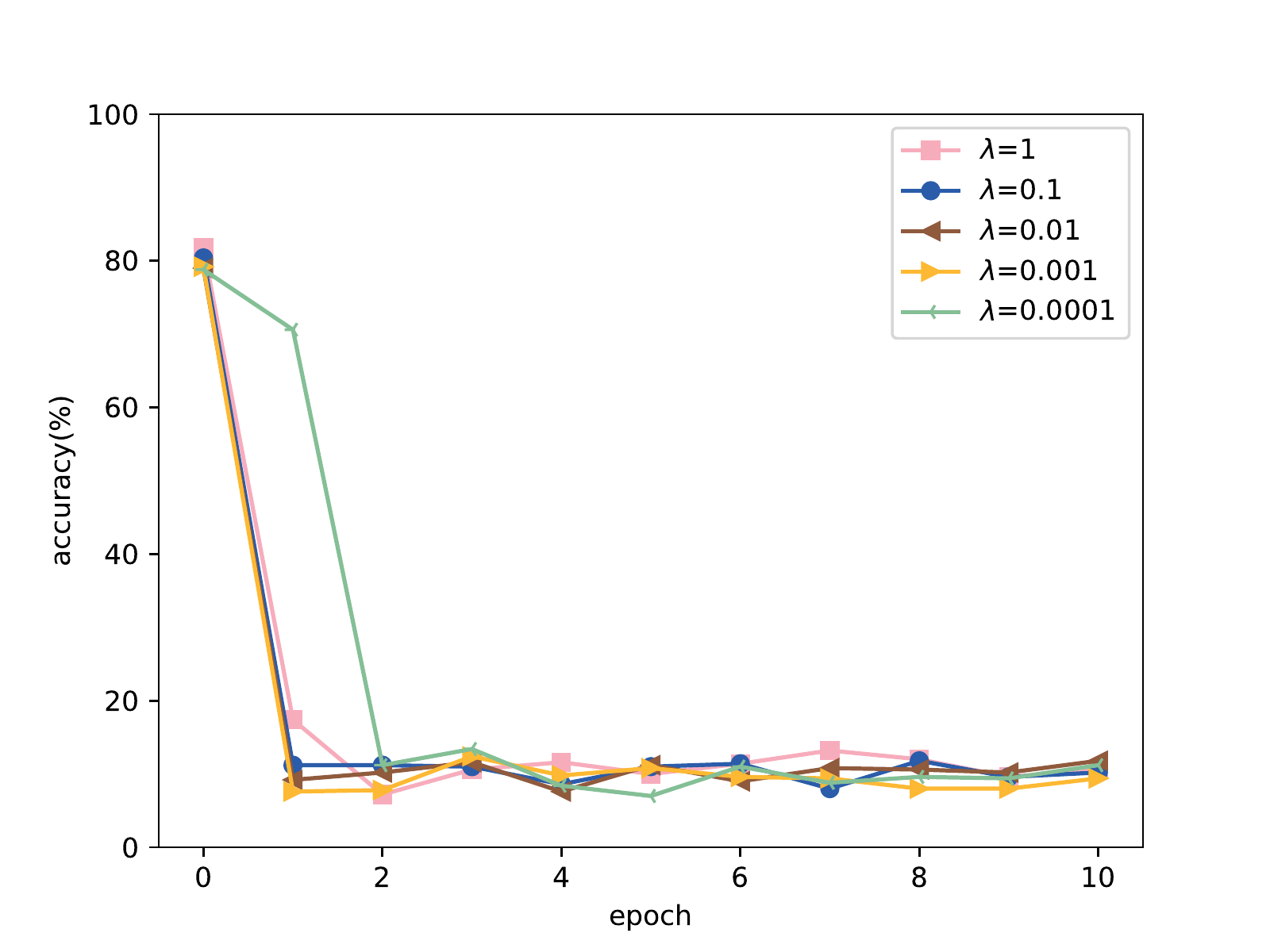}
		\caption{Influence of different of $\lambda$ values on the accuracy of cifar10.R dataset after unlearning (left)
			Influence of different values of $\lambda$ on the accuracy of backdoor attack after cifar10.R dataset unlearning (right)}
		\label{fig:minipage4}
	\end{figure*}
	
	\subsection{The influence of the number of training data in reference model}
	
\begin{figure*}[ht]
		\centering
		\includegraphics[width=7cm]{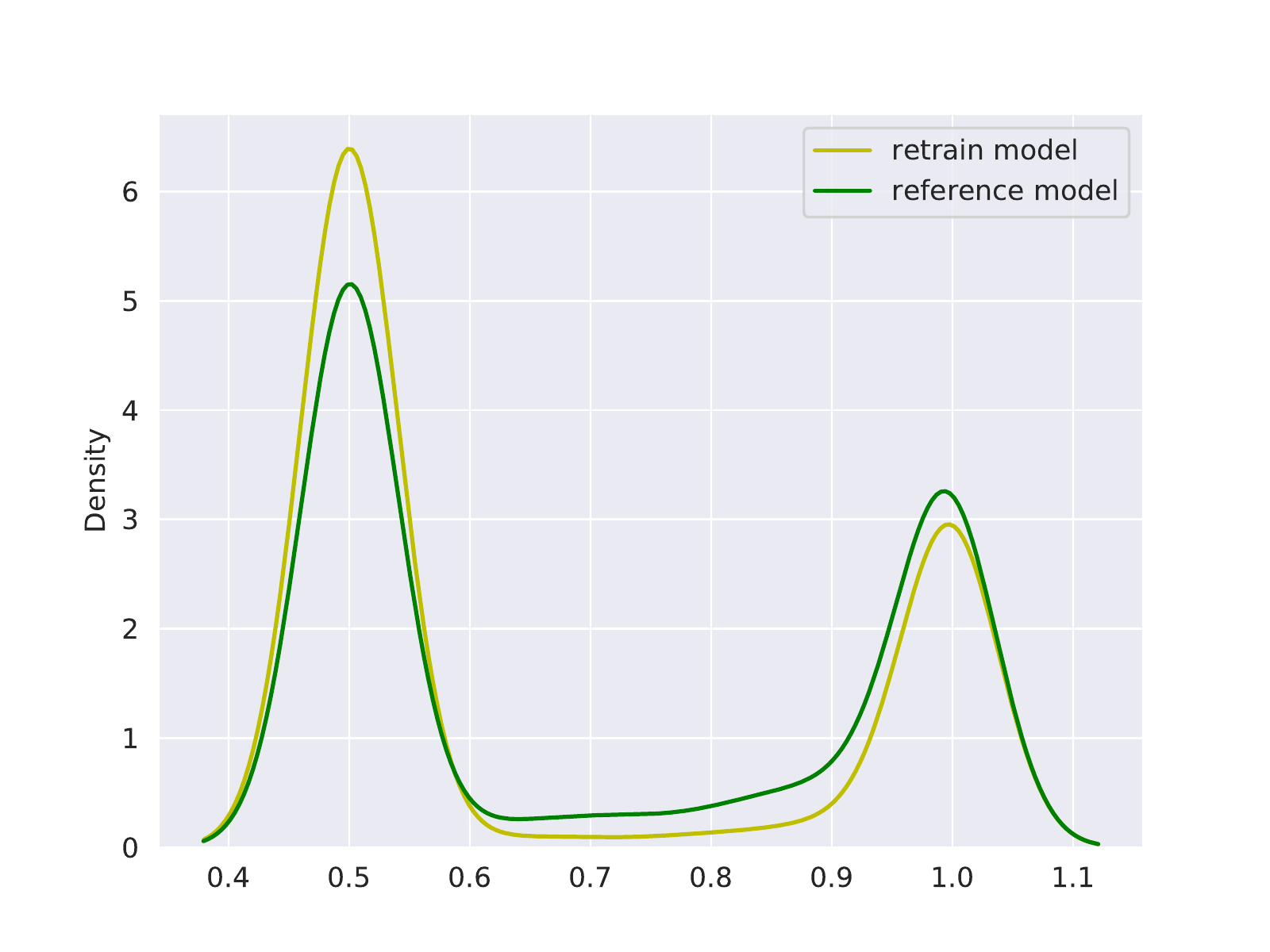}
		\hspace{0.5in}
		\includegraphics[width=7cm]{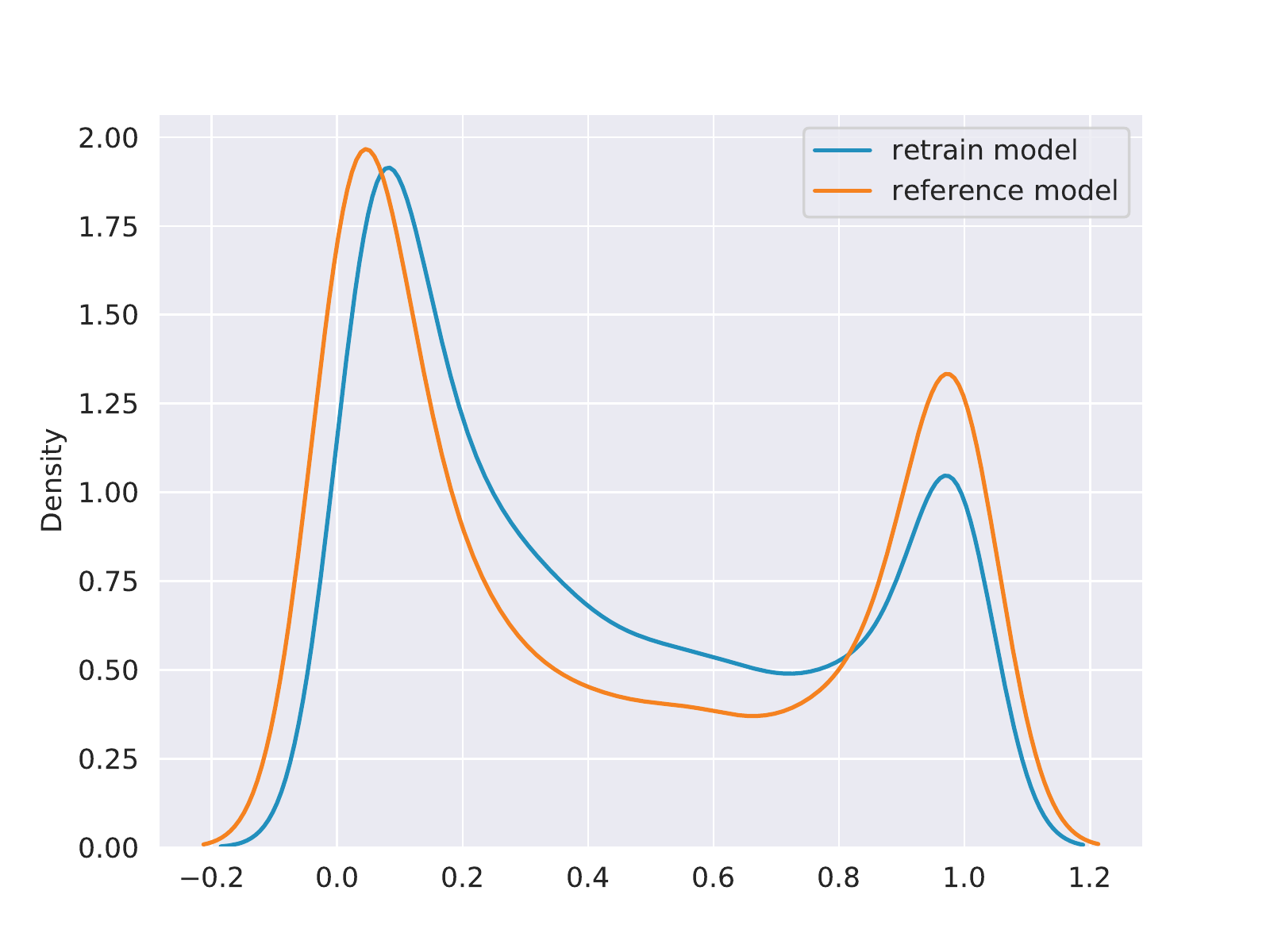}
		\caption{Comparison between the output distribution of the mnist dataset in the retraining model and that in the reference model (left)
			Comparison between the output distribution of the cifar10.R dataset in the retraining model and that in the reference model (right)
		}
		\label{fig:dis_cifar}
	\end{figure*}

	As mentioned in Chapter 1, in order to eliminate the influence of forgotten data on the target model, we want to make the target model iteratively close to the direction of the model retrained by $D_r$. Considering the time cost, we select a subset $D_s$ from $D_r$ And generate the reference model $M_0$ by $D_s$ training. Iterate the target model in the direction of the reference model.As shown in figure \ref{fig:dis_cifar} on the left, the posterior distribution of the retraining model almost coincides with that of the reference model on the MNIST dataset. The peaks are clustered around 0.5 and 1, because the output values of the model Lenet used in the MNIST dataset are positive through the relu function. The probability of 0 becomes 0.5 after being output through the sigmoid function, and 1 remains unchanged. As shown in figure \ref{fig:dis_cifar} on the right, the posterior distribution of the retrained model is also very close to that of the reference model on the CIFAR10.R dataset.This shows that the training direction of the reference model is consistent with that of the retraining model, so it is a correct choice to take the reference model as the reference object that we forget and reduce the time cost.

	For different datasets, we take subset $D_s$ from the remaining data respectively to train the reference model $M_0$, and the default Settings of $M_0$ are shown in the table \ref{fig:default_M}. The structure of initial model of each $M_0$ is the same as that of model $M$ before unlearning. As table \ref{fig:default_M}. As shown in the Mnist dataset, we used 1/10 of the total amount of the total training set $D$, namely 6000 samples, to train M0.  After the end of iteration, namely 50 epochs, the training accuracy could reach 85.52\%, and the test accuracy was 85.45\%.It took 42.81s to train the reference model. The same goes for other datasets.

	\renewcommand\arraystretch{1.5}
	\begin{table}[!h] 
		\centering
		\caption{Training information, training time and accuracy of reference model}
		\label{fig:default_M} 
		\scalebox{0.6}{
			\begin{tabular}{p{3cm}cccccc} 
				\hline 
				dataset & number of instance & Acc.train(\%) &  Acc.test(\%) & Time(s)\\
				\hline
				Mnist		  & 6000  & 87.40 & 87.05 & 42.81\\
				Fashion-mnist & 6000  & 79.17 &	78.38 & 67.59\\
				SVHN		  & 7325  & 79.45 & 79.90 & 140.29\\
				Cifar10.R	  & 5000  & 69.90 & 68.40 &	143.70\\
				Cifar10.V	  & 5000  & 66.98 & 65.56 & 141.95\\
				\hline
		\end{tabular}}   
	\end{table}
	
	As table \ref{fig:mnist_M} shown, mnist dataset is taken as an example, and the sample numbers are 2000, 4000, 6000, 10000 and 15000 respectively as the training data quantity of the reference model.When the value is 6000, the training accuracy and testing accuracy obtained after the convergence of the reference model are 86.35\% and 86.26\% respectively, and the training time is 42.81s. Using the current model as the reference model for unlearning, the accuracy of the forgotten model is 98.27\%. Testing the current forgotten model has a 19\% accuracy rate for backdoor attacks.The fashion-Mnist dataset has the same effect.

	\renewcommand\arraystretch{1.5}
	\begin{table*}[!h] 
		\centering
		\caption{Influence of training sample number of reference model on MNIST}
		\label{fig:mnist_M}   
		\scalebox{0.8}{
			\begin{tabular}{p{3.2cm}ccccccc} 
				\hline 
				number of instance &  Acc.test(\%) & unlearn Acc.test(\%) & Membership
				attack accuracy(\%) & Time(s)\\
				\hline
				2000    & 83.05 & 98.08 & 50.00 & 24.07\\
				4000    & 86.26 & 98.53 & 51.33 & 30.55\\
				6000    & 87.05 & 98.27 & 49.08 & 42.81\\
				10000   & 88.62 & 98.56 & 51.75 & 62.78\\
				15000   & 97.53 & 98.48 & 50.42 & 90.33\\
				\hline
		\end{tabular}} 
	\end{table*}
	
	\begin{figure*}[ht]
		\centering
		\includegraphics[width=7cm]{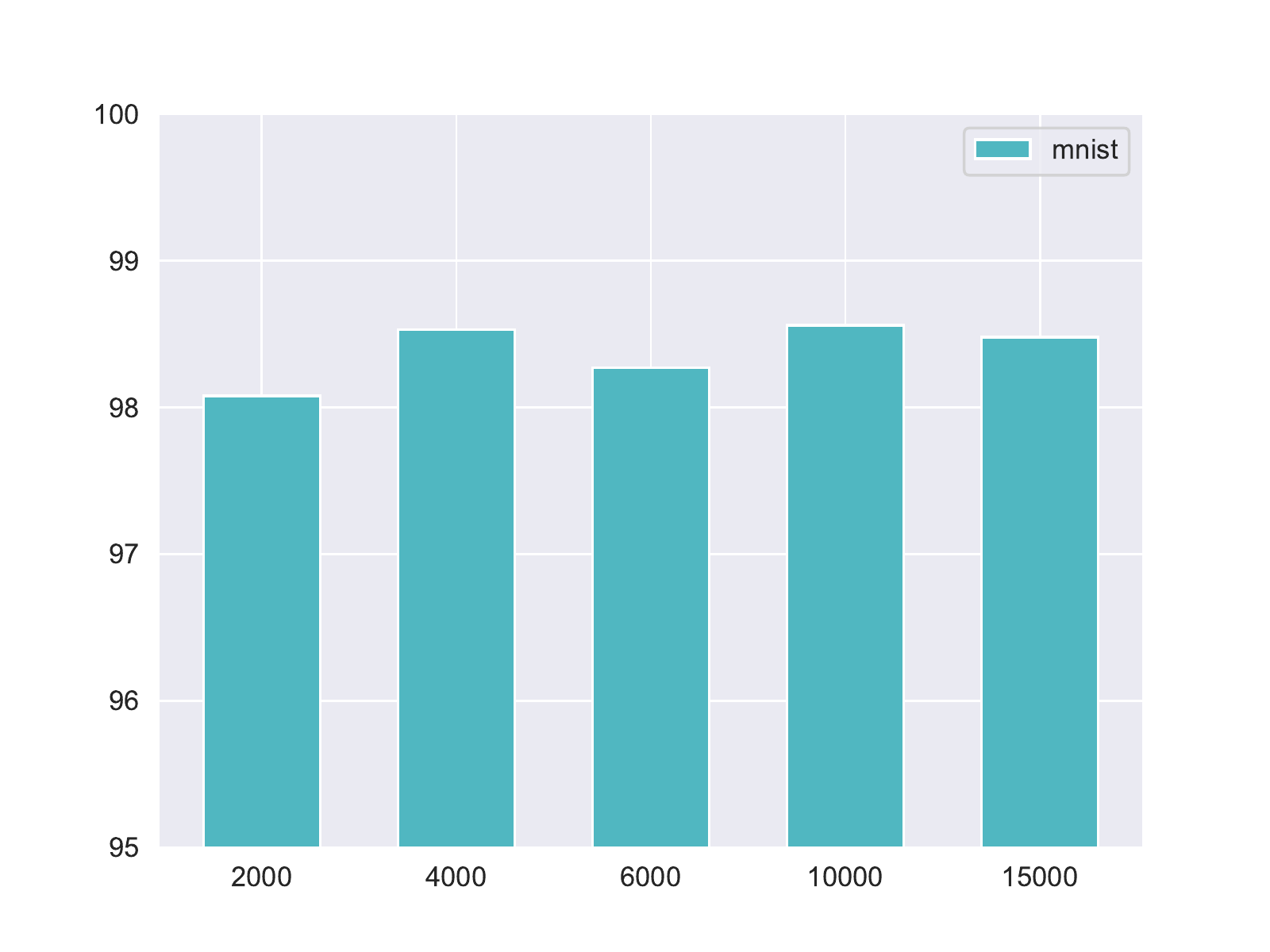}
		\label{fig:M_mnist}
		\hspace{0.5in}
		\includegraphics[width=7cm]{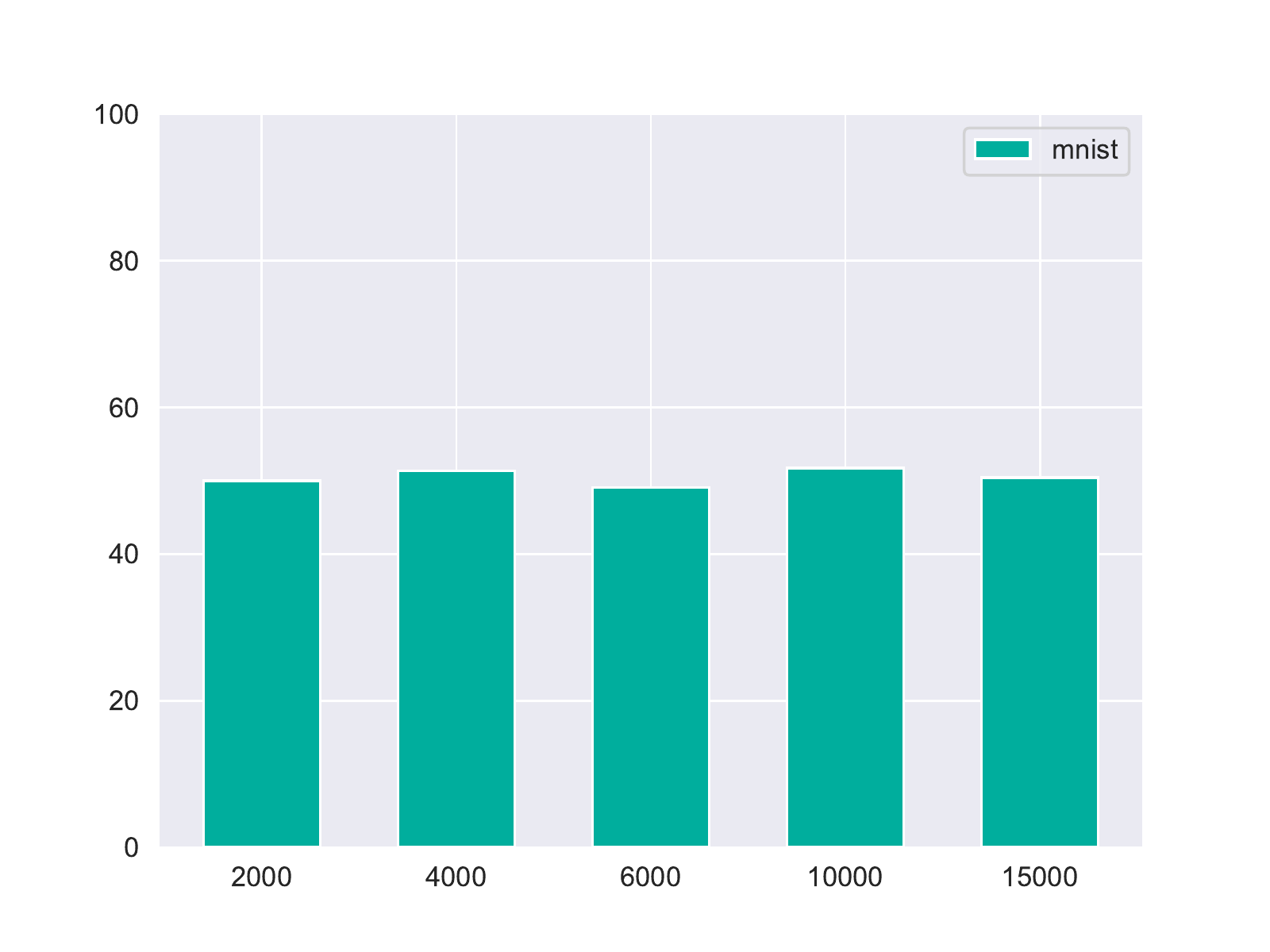}
		\caption{Effect of MNIST dataset reference model training data size on the accuracy of target model after unlearning (left)
			Effect of MNIST dataset reference model training data size on inference attack accuracy of target model members after unlearning (right)
		}
	\end{figure*}

	\renewcommand\arraystretch{1.5}
	\begin{table*}[!h] 
		\centering
		\caption{Influence of training sample number of reference model on cifar10.R}
		\label{fig:cifar_M}    
		\scalebox{0.8}{
			\begin{tabular}{p{3.2cm}cccccc} 
				\hline 
				number of instance  & Acc.train(\%) &  Acc.test(\%) & unlearn Acc.test(\%) & Membership attack accuracy(\%) & Time(s)\\
				\hline
				3000   & 56.09 & 54.57 & 87.34 & 49.90 & 41.62\\
				5000   & 69.90 & 68.40 & 85.85 & 52.30 & 143.70\\
				7500   & 76.00 & 73.76 & 87.90 & 52.70 & 821.07\\
				10000  & 78.53 & 76.91 & 87.64 & 51.00 & 1233.21\\
				15000  & 82.48 & 80.51 & 90.24 & 53.20 & 1574.09\\
				\hline
		\end{tabular}}
	\end{table*}

	\begin{figure*}[ht]
		\centering
		\includegraphics[width=7cm]{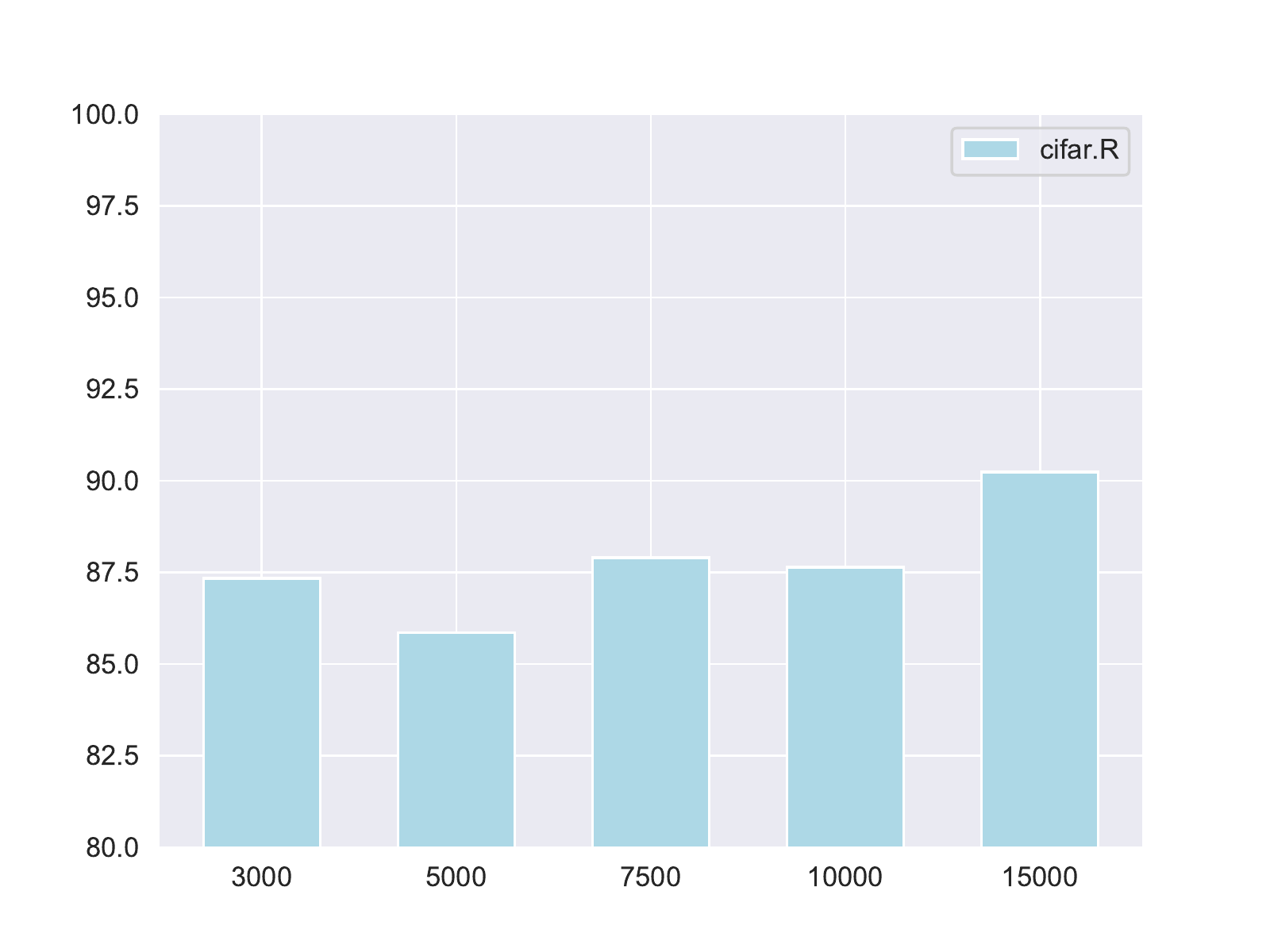}
		\hspace{0.5in}
		\includegraphics[width=7cm]{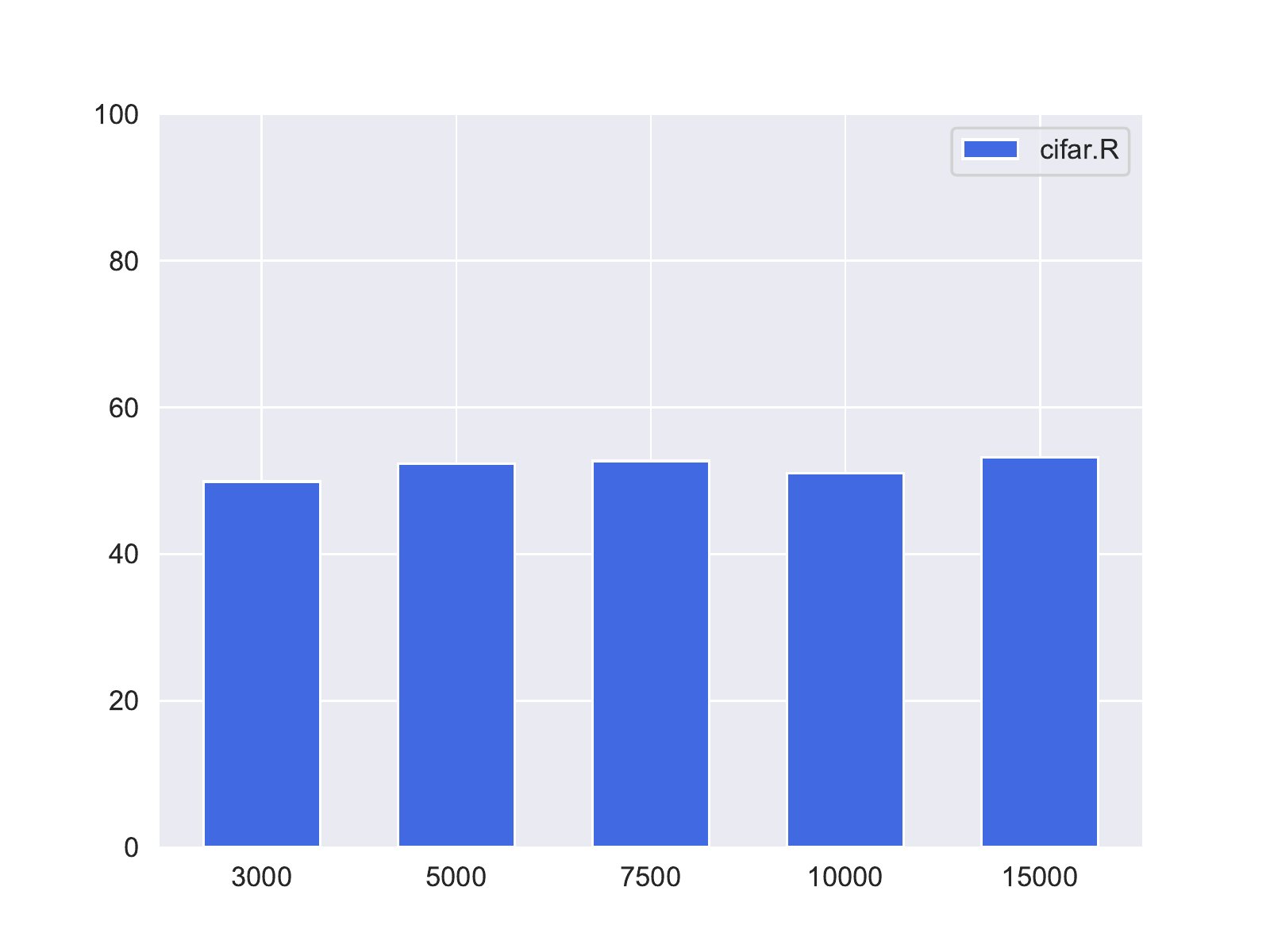}
		\caption{Effect of training data size of cifar10 dataset reference model on the accuracy of target model after unlearning (left)
			Effect of training data size of cifar10 dataset reference model on inference attack accuracy of target model members after unlearning (right)}
		\label{fig:M_cifar}
	\end{figure*}

	As table \ref{fig:cifar_M} shown, for the Cifar.R dataset, the sample numbers are 3000, 5000, 7500, 10000 and 15000 respectively as the training data quantity of the reference model.When the value is 5000, the training accuracy and testing accuracy obtained by referring to model convergence are 69.90\% and 68.40\% respectively, and the training time is 143.7s. Using the current model as the reference model for unlearning, the accuracy of the forgotten model is 85.85\%.The accuracy of member inference attack is 52.30\%. The Cifar.V dataset has the same effect. As shown in figure \ref{fig:M_cifar}, when the reference model training sets of different sizes are taken, there is no obvious influence trend on the model accuracy after the unlearning.

	\subsection{Distribution before and after unlearning}

	\begin{figure*}[ht]
		\centering
		\includegraphics[width=7cm]{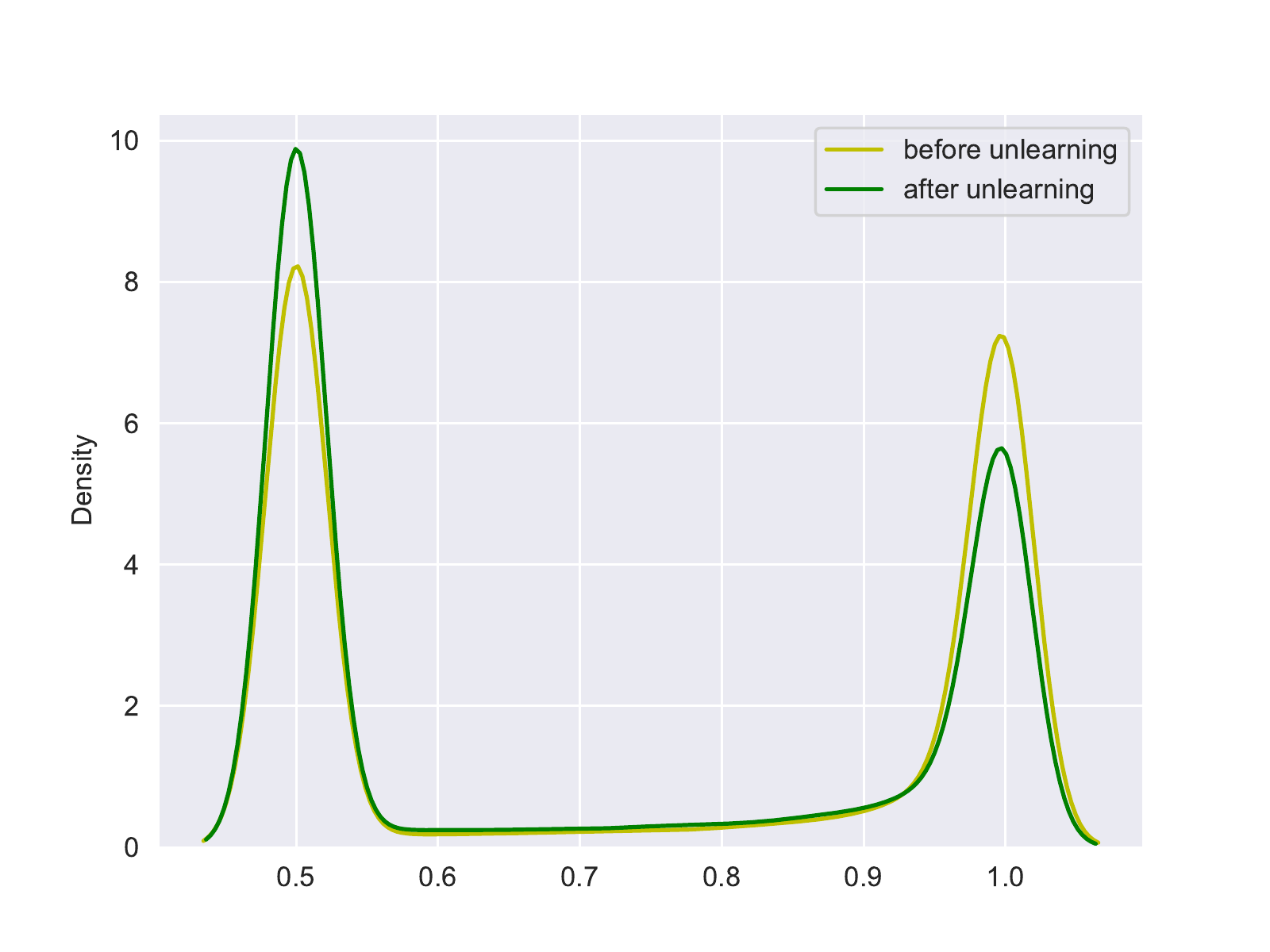}
		\hspace{0.5in}
		\includegraphics[width=7cm]{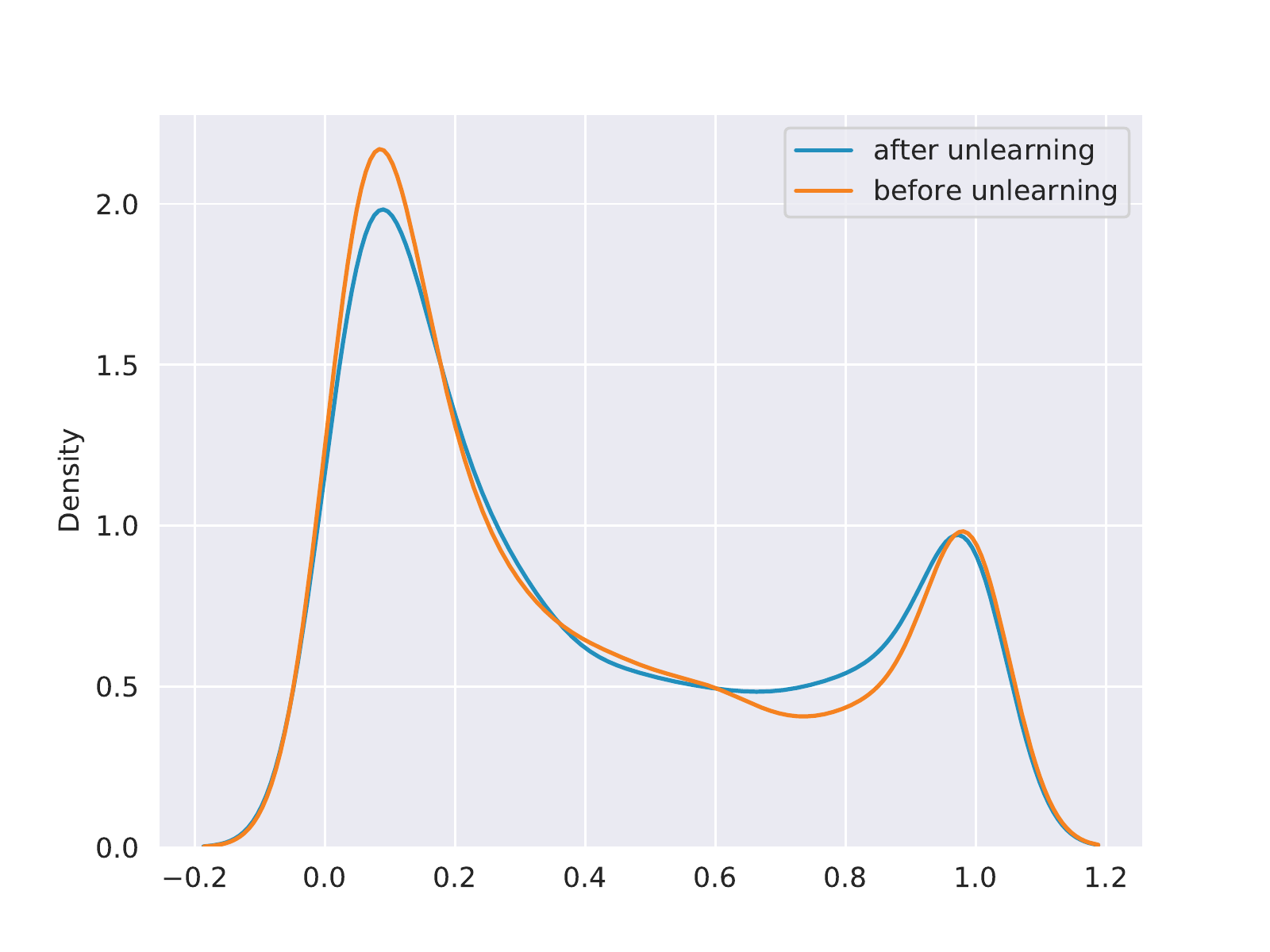}
		\caption{Comparison of output distribution of MNIST dataset before and after unlearning (left)
			Comparison of output distribution of cifar. R dataset before and after unlearning (right)
		}
		\label{fig:after_cifar}
	\end{figure*}
	
	Posterior distribution of the model before and after unlearning: In order to verify that machine unlearning by our method does not affect the performance of the model itself, the posterior distribution of different datasets before and after unlearning and the weight distribution of the target model before and after unlearning are counted. As shown in figure \ref{fig:after_cifar} on the left, the mnist dataset's output distributions before (yellow line) and after (green line) unlearning almost coincide. As Figure \ref{fig:after_cifar} on the right shown, the output distributions of cifar10.R dataset before unlearning (blue line) and after unlearning (red line) also almost coincide, which means that the model before and after unlearning of the target model can be almost unaffected by using our method.

	\section{Related work}

	Existing work includes Cao \cite{2015Towards} proposing a method of using statistical queries to obtain dataset features rather than training them directly, thus unlearning them in less time than retraining them. However, the statistical query method is only applicable to non-adaptive machine learning models (late training does not depend on early training), and it is difficult to achieve unlearning effect on neural networks. The dataset is divided into several parts, each part is trained into a separate sub-model and stored, and the overall model is trained through incremental learning. To forget a sample, retraining starts with the first intermediate model that contains the sample's contribution. However, this method reduces training time, but consumes a lot of storage space. Liu \cite{liu2020federated} proposed data unlearning based on federated learning framework is to save the update parameters of each round in the normal training model aggregation stage, and then reduce the number of iterations of client training when deleting the forgotten data and retraining. Model aggregation combines the parameters of the current client and the updated parameters saved during the previous training to construct the server model.This approach reduces training time, but also takes up more storage space because saving parameters. Due to the need to save updated parameters, the training process of the target model is modified, and the saved parameters themselves carry information to be forgotten, which theoretically cannot guarantee the complete unlearning. 
	Ginart et al. 's \cite{ginart2019making} study the problem of retraining model after removing data to be forgotten, especially applied to k-means algorithm, which is difficult to be applied to supervised learning. 
	Izzo et al. \cite{izzo2021approximate} mainly studies the unlearning of data on linear computation, and completes linear scaling on data dimensions. 
	\cite {baumhauer2020machine} proposes a linear transformation method applied to the classifier to re-scale the logarithmic probability predicted by the classifier, but does not eliminate the contribution of relevant information to the model. 
	\cite{guo2019certified} based on differentiable convex functions (e.g., logistic regressors), a unlearning method is proposed that uses Newton's method to remove weight information and differential privacy to mask unlearning residuals. 
	The study of unlearning methods in Bayesian models \cite{nguyen2020variational} introduces new techniques for adjusting likelihood and reverse KL.  \cite{neel2020descent} uses gradient descent to remove data from a convex function. 
	In comparison, our method is highly applicable to neural networks. A small perturbation is made on the parameters of the original target model, the KL divergence of the posterior probability of the target model and the reference model is taken as the loss, and the precision penalty term is added to avoid catastrophic unlearning. The training process does not change the target model and does not occupy additional storage space. It has relatively small cost in time and space and almost no loss of model accuracy.

	Our work is also inspired by membership inference attack and backdoor attacks to evaluate the effects of unlearning. Membership inference attack \cite{2017Membership}, \cite{salem2018ml}, \cite{wang2019miasec}, \cite{jia2019memguard}, \cite{truex2019demystifying} answer the question of whether to use a specific sample to train a machine learning model. 
	The method proposed by Song et al. \cite{song2019auditing} can judge whether a user data participates in the training of machine learning model. Multiple shadow models and data need to be used, resulting in a large cost of time and space. 
	Later, \cite{salem2018ml} proposed that it is unnecessary to train multiple shadow models, and the performance of member reasoning attack is not affected by the number of shadow models. 
	\cite{sommer2020towards}, \cite{sablayrolles2020radioactive} believes that there are few methods to evaluate machine unlearning, and member reasoning attack is very suitable to be used as a method to evaluate machine unlearning. In this paper, we also apply member inference attack to evaluate the amount of residual information after machine unlearning.
	
	Backdoor attack is one of the evaluation indexes in this paper. 
	\cite{2017BadNets} selects a subset of the training set, places a back door on this part of the samples, and takes a small area to set the color or brightness to zero. The corresponding labels of this part of data are uniformly modified to a specific label, and the target model is trained together with other original data. Finally, the target model learns to recognize the originally set specific label according to the back door.  
	Tang et al. \cite{tang2020embarrassingly} designed a method without training in advance and changing the sample itself. Just insert a small Trojan horse module into the model without modifying the parameters in the original model.
	The research work of Liu et al. \cite{liu2020reflection} considers that the modification of training data and its labels can be easily detected by input filtering defense strategies, and their scheme is more natural and does not need to modify labels. Through the mathematical modeling of the physical reflection model, the radiation image of the object is used as the back door implantation model.

	\section{Conclusion}
	Machine unlearning needs to completely eliminate the contribution of information to be forgotten from the model. This paper proposes a method to solve this problem, which does not need to retrain the model, does not consume a lot of time and space cost, and is highly suitable for the neural network model. Based on a reference model without unlearning information, the target model iteratively approaches in this direction until the contribution of unlearning information is completely eliminated. Our method has progressive significance in privacy protection, and can also be applied to many interesting directions, including rapid adjustment of trainable parameters of the model, malicious erasure model related contributions, model defense, and data transaction and sharing. Our method has good applicability for neural networks, and may have limitations for other machine learning models. We hope to continue to break through and extend in future work. Taking our work as a stepping stone, we hope that more people can pay attention to privacy security. Future research can further go deep into the field of privacy security and provide broader methods to protect private user data.

	\bibliographystyle{IEEEtran}
	\bibliography{reference}

\end{document}